\documentclass[pmlr]{jmlr}

\usepackage{multirow}
\usepackage{rotating}
\usepackage{longtable}

 \usepackage{booktabs}
 \usepackage{mathtools}
 
\usepackage[load-configurations=version-1]{siunitx} 
\DeclareMathOperator*{\argmax}{arg\,max}
\usepackage{array}

\makeatletter
\def\set@curr@file#1{\def\@curr@file{#1}} 
\makeatother

\theorembodyfont{\upshape}
\theoremheaderfont{\scshape}
\theorempostheader{:}
\theoremsep{\newline}

\jmlrvolume{182}
\jmlryear{2022}
\jmlrworkshop{Machine Learning for Healthcare}

\title[Anomaly Detection in Echocardiograms]{Anomaly Detection in Echocardiograms with Dynamic Variational Trajectory Models}
\usepackage{authblk}
\author[1]{Alain Ryser}
\author[1]{Laura Manduchi}
\author[1]{Fabian Laumer}
\author[2]{Holger Michel}
\author[2]{Sven Wellmann}
\author[1]{Julia E. Vogt}

\affil[1]{Department of Computer Science, ETH Zurich}
\affil[2]{Department of Neonatology, University Children’s Hospital Regensburg (KUNO), University of Regensburg, Germany}

\begin{document}

\maketitle

\begin{abstract}
We propose a novel anomaly detection method for echocardiogram videos.
The introduced method takes advantage of the periodic nature of the heart cycle to learn three variants of a \emph{variational latent trajectory} model (TVAE).
While the first two variants (TVAE-C and TVAE-R) model strict periodic movements of the heart, the third (TVAE-S) is more general and allows shifts in the spatial representation throughout the video.
All models are trained on the healthy samples of a novel in-house dataset of infant echocardiogram videos consisting of multiple chamber views to learn a normative prior of the healthy population.
During inference, maximum a posteriori (MAP) based anomaly detection is performed to detect out-of-distribution samples in our dataset.
The proposed method reliably identifies severe congenital heart defects, such as Ebstein's Anomaly or Shone-complex.
Moreover, it achieves superior performance over MAP-based anomaly detection with standard variational autoencoders when detecting pulmonary hypertension and right ventricular dilation.
Finally, we demonstrate that the proposed method enables interpretable explanations of its output through heatmaps highlighting the regions corresponding to anomalous heart structures.
\end{abstract}

\section{Introduction}

Congenital heart defects (CHDs) account for about 28\% of all congenital defects worldwide \citep{van2011birth}. 
CHDs manifest in several different heart diseases with various degrees of frequency and severity and are usually diagnosed primarily with echocardiography. 
Echocardiography is one of the most common non-invasive screening tools due to the rapid data acquisition, low cost, portability, and measurement without ionizing radiation. 
Early screening of heart defects in newborns is crucial to ensure the long-term health of the patient \citep{buskens1996efficacy,singh2016fetal,van2016prenatal}. 
However, due to the subtlety of various heart defects and the inherently noisy nature of echocardiogram video (echo) data, a thorough examination of the heart and the diagnosis of CHD remains challenging and time-consuming, raising the need for an automated approach. 
Still, collecting real-world datasets from large populations to apply state-of-the-art supervised deep learning methods is often infeasible.
The reason is that many CHDs like Ebstein's Anomaly, Shone-complex, or complete atrioventricular septal defect (cAVSD) rarely occur, making the dataset extremely imbalanced. 
On the other hand, we have access to an abundance of echos from healthy infant hearts generated during standard screening procedures, often performed on infants shortly after birth. 
In this work, we leverage the healthy population and propose a novel anomaly detection method to identify a variety of CHDs. 
The proposed approach learns a structured normative prior of healthy newborn echos using a periodic variational latent trajectory model. 
At test time, the method can detect out-of-distribution samples corresponding to CHDs. The advantage of this approach is that the model is trained purely on healthy samples, eliminating the need to collect large amounts of often rarely occurring CHDs.

In anomaly detection, we assume that all data is drawn from a space $\mathcal{X}$ with some probability density $p_H$.
We define anomalies as samples drawn from low probability regions of $\mathcal{X}$ under $p_H$.
More formally, the space of anomalies $\mathcal{A}\subset\mathcal{X}$ under density $p_H$ and anomaly threshold $\tau\geq0$ is defined by
$$
\mathcal{A}=\{x\in\mathcal{X}; p_H(x)\leq\tau\}
$$
Note that $\tau$ is a task-specific measure, as the definition of anomaly can vary drastically over different problem settings.
Consequently, most anomaly detection algorithms assign anomaly scores rather than discriminating between normal and anomalous samples.

In this work, we focus on reconstruction-based approaches, which encompass some of the most widespread methods for anomaly detection \citep{chalapathy2019deep, ruff2021unifying,pang2021deep}.
This family of methods aims to learn generative models that can reconstruct normal samples well but decrease in performance for anomalous inputs. 
A given measure $\alpha_f(x)$ that quantifies the reconstruction quality achieved by model $f$ when given sample $x$ can then be interpreted as the anomaly score of $x$. 
The models are commonly trained on healthy samples, and during inference, an anomalous sample $x_a$ is assumed to get projected into the learned normal latent space.
This leads to high reconstruction errors, resulting in high anomaly scores $\alpha_f(x_a)$.
More recently, \cite{chen2020unsupervised} proposed a variation of the reconstruction-based approach that allows us to incorporate prior knowledge on anomalies during inference by detecting anomalies using a \emph{maximum a posteriori} (MAP) based approach.
However, this approach requires an estimate of the log-likelihood, which restricts model choice to generative models such as \emph{variational autoencoders} (VAE \citet{kingma2013auto}).

Although various generative architectures have been proposed in the literature, little effort has been directed toward echocardiogram videos. 
One exception is the work of \citet{laumer2020deepheartbeat}, where the authors introduced a model that specifically targets the periodicity of heartbeats for ejection fraction prediction and arrhythmia classification. 
However, the model enforces somewhat restrictive assumptions on the heart dynamics and is purely deterministic. 
In contrast, we propose a variational latent trajectory model that overcomes the simplistic assumptions of previous approaches and learns a distribution over dynamic trajectories, enabling the detection of different types of CHDs in echocardiograms using the MAP approach. 
Furthermore, the proposed algorithm produces incredibly accurate video reconstructions from surprisingly few latent factors.

To allow the community to reuse our methods in future work, we provide code for the TVAE and pretrained models used in our experiments on Github\footnote{\url{https://github.com/alain-ryser/tvae}}.

\subsection*{Generalizable Insights about Machine Learning in the Context of Healthcare}
This work introduces a novel generative model designed explicitly for echocardiogram videos.
By leveraging key assumptions on this data modality, our model allows a flexible encoding of videos in a small number of latent dimensions from which accurate reconstructions can be retrieved.
We demonstrate how this method outperforms previous approaches on an anomaly detection task using a new in-house echo video dataset containing samples of newborns and infants with various forms of congenital heart defects. 
Our model learns an accurate normative prior on healthy echo data and then performs maximum a posteriori-based anomaly detection to detect CHDs.
Additionally, we demonstrate how our model produces interpretable outputs by showcasing decision heatmaps that highlight regions that drive anomaly scores.
To summarize, the contributions of this paper are the following:
\begin{enumerate}
    \item We propose a novel variational latent trajectory model (TVAE) for reconstruction-based anomaly detection on echocardiogram videos.
    \item We perform extensive evaluation of the proposed method on the challenging task of CHD detection in a real-world dataset.
    \item We complement our predictions with decision heatmaps, highlighting the echocardiogram regions corresponding to anomalous heart structures.
\end{enumerate}

\section{Related Work}

The rapid data acquisition, the high observer variation in their interpretation, and the non-invasive technology have made echocardiography a suitable data modality for many machine learning algorithms. In recent years, a variety of algorithms for \emph{Segmentation} \citep{dong2016left,moradi2019mfp,leclerc2019deep}, \emph{View Classification} \citep{gao2017fused, vaseli2019designing} or \emph{Disease Prediction} \citep{madani2018deep,kwon2019deep} have been proposed. However, their performance often relies on the assumption that a large \emph{labeled} dataset can be collected. This assumption does not hold for rare diseases, where the amount of collected data is often too scarce to train a supervised algorithm. 
 Hence, reconstruction-based anomaly detection algorithms could be used in such a setting, but their application to echocardiography is, to the best of our knowledge, left unexplored.

Previous work on reconstruction-based anomaly detection are often based on generative models, such as \emph{autoencoders} (AE) \citep{chen2017outlier,principi2017acoustic,chen2018unsupervised,pawlowski2018unsupervised} or \emph{variational autoencoders} (VAE \citet{kingma2013auto}) \citep{an2015variational,park2018multimodal,xu2018unsupervised,cerri2019variational,you2019unsupervised}. 
Their application to the medical domain is mostly limited to disease detection in MRI, \citep{chen2018unsupervised,baur2018deep,baur2020steganomaly,chen2020unsupervised,baur2021modeling,pinaya2021unsupervised} where anomalies are often easily detectable as they are clearly defined by regions of tissue that contain lesions.
On the other hand, pathologies of CHDs in echos are largely heterogeneous and can usually not be described by unique structural differences from healthy echos.
Identifying them is often challenging, as they can be caused by small perturbations of ventricles (ventricular dilation) or subtle malfunctions like pressure differences between chambers in certain phases of the cardiac cycle (pulmonary hypertension).
Detecting certain CHDs thus requires the inclusion of temporal structures in addition to the spatial information leveraged in MRI anomaly detection.

Different extensions to AE/VAE have been proposed to perform reconstruction-based anomaly detection methods on video data \citep{xu2015learning,hasan2016learning,yan2018abnormal}. However, these methods are often mainly designed for abnormal event detection, where anomalies can arise and disappear throughout the video.
On the other hand, we are interested in whether a given video represents a healthy or anomalous heart.
Another method for video anomaly detection is \emph{future frame prediction} \citep{liu2018future}.
This approach trains models to predict a video frame from one or more previous ones.
During inference, it is then assumed that such a model achieves better performance on normal than on anomalous frames.
Recently, \cite{yu2020cloze} proposed a method that combines reconstruction and future frame prediction-based approaches in one framework.
Though achieving good performance on videos with varying scenes, future frame prediction does not seem suitable for echos as just returning any input frame will always lead to good prediction scores due to the periodic nature of the cardiac cycle.
An entirely different approach to anomaly detection is given by \emph{One-Class Classification} \citep{moya1996network}. 
In contrast to the previous approaches, the latter relies on discriminating anomalies from normal samples instead of assigning an anomaly score. 
This is usually achieved by learning a high-dimensional manifold that encloses most or all normal data. The surface of this manifold then serves as a decision boundary that discriminates anomalies from normal samples.
One of the more prominent methods of that family is the so-called \emph{Support Vector Data Description} (SVDD) \citep{tax2004support} model.
The SVDD learns parameters of a hypersphere that encloses the training data.
Similar to SVMs, it provides a way to introduce some slack into the estimation process, allowing certain normal samples to lie outside the decision boundary. 
A similar approach is given by the \emph{One-Class SVMs}  (OC-SVM) \citep{scholkopf2001estimating}, where anomalies are discriminated from normal samples by learning a hyperplane instead of a hypersphere. 
Like with SVMs, the expressivity of SVDD and OC-SVM can be drastically improved by introducing kernelized versions \citep{ratsch2002constructing,ghasemi2012bayesian,dufrenois2014one,gautam2019localized}.
More recently, deep neural networks have been proposed to perform anomaly detection based on similar principles \citep{sabokrou2018adversarially,ruff2018deep,ruff2020rethinking,ghafoori2020deep}.
While conceptually interesting, One-Class Classification methods often require large amounts of data to work accurately, making them unsuitable in many clinical applications.

\section{Methods}\label{sec:methods}
In this work, we propose a probabilistic latent trajectory model to perform reconstruction-based anomaly detection on echocardiogram videos.
To that end, we take inspiration from latent trajectory models \citep{Louis2019, laumer2020deepheartbeat} and introduce a variational autoencoder that learns a structured normative distribution of the heart's shape and dynamic. In particular, the model encodes the echos into stochastic trajectories in the latent space of a VAE, enabling us to accurately generate high-quality reconstructions while maintaining a low dimensional latent bottleneck.
The learned approximate distribution of healthy hearts allows us to detect anomalies post-hoc using a maximum a posteriori (MAP) approach \citep{chen2020unsupervised}.
High-quality normative reconstructions and informative latent representations are essential for correctly detecting out-of-distribution echos.

\subsection{Latent Trajectory Model}\label{subsec:traj_models}
\begin{figure*}[t!]
    \centering
    \includegraphics[width=\textwidth]{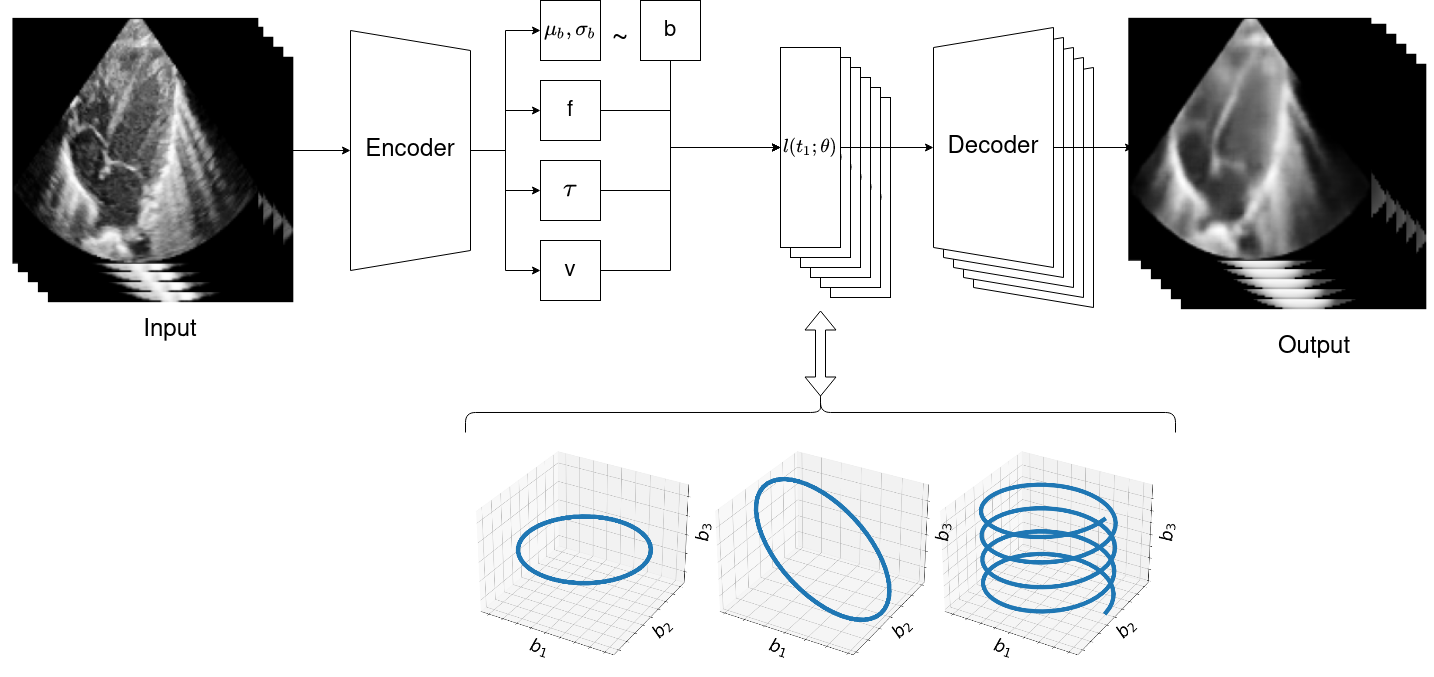}
    \caption{Overview of the model architecture with $\vec{\ell}_{circular}$ (left), $\vec{\ell}_{rot}$ (middle) and $\vec{\ell}_{spiral}$~(right).}
    \label{fig:trajectories}
\end{figure*}
The latent trajectory model \citep{laumer2020deepheartbeat} is an autoencoder that is designed to learn latent representations from periodic sequences of the heart, i.e. echos in this case. The main idea is to capture the \emph{periodic nature} of the observed data by learning an encoder $\phi$ that maps an echo $X \coloneqq (\vec{x}^{(j)}, t^{(j)})_{j=1}^T$ with frames $\vec{x}^{(j)} \in \mathbb{R}^{w \times h}$ at time points $t^{(j)}$ to a prototypical function $\vec{\ell}_{circular}(t;\phi(X))$ whose parameters contain information about the heart's shape and dynamic.
The decoder $\psi$ reconstructs the original video frame by frame from the latent embedding $\vec{\ell}_{circular}$ via
$$
\Tilde{\vec{x}}^{(j)}=\psi(\vec{\ell}_{circular}(t^{(j)};\phi(X)))
$$
Here, $\vec{\ell}_{circular}$ corresponds to the following cyclic
trajectory:
$$\vec{\ell}_{circular}(t;f,\omega,\vec{b})=\begin{pmatrix}
\cos(2\pi f t - \omega) + b_1\\
\sin(2\pi f t - \omega) + b_2\\
b_3 \\
\vdots \\
b_d
\end{pmatrix},
$$
where the frequency parameter, $f > 0$, corresponds to the number of cycles per time unit, and the offset parameter $\omega\in[0,2\pi]$ allows the sequence to start at an arbitrary point within the (cardiac) cycle. The parameter $\vec{b}\in \mathbb{R}^d$ characterizes the \textit{spatial information} of the signal. 
See Figure~\ref{fig:trajectories} a) for an illustration of $\ell_{circular}$.
This model thus describes a simple tool to learn the disentanglement of temporal components ($f$, $\omega$) from a common spatial representation ($\vec{b}$) for a given echo.
On the other hand, the assumptions made may be too simplistic to result in good reconstructions.
We will address this issue in the following sections.

\subsection{Dynamic Trajectories}
The above formulation, $\vec{\ell}_{circular}$, allows modeling time-related information only through the first two latent dimensions, thereby limiting the amount of time-dependent information that can be encoded in the latent space. 
The reduced flexibility results in insufficient reconstruction quality, impairing the reconstruction-based anomaly detection performance.
To circumvent this problem, we distribute time-dependent components over each dimension of the latent space while retaining the periodicity.
We thus define the rotated trajectory function $\vec{\ell}_{rot}$ as

$$
\vec{\ell}_{rot}(t;f,\omega,\vec{b})=
\begin{pmatrix}
    \cos(2\pi f t - \omega) - \sin(2\pi f t - \omega) + b^{(1)}\\
    \cos(2\pi f t - \omega) + \sin(2\pi f t - \omega) + b^{(2)}\\
    \vdots \\
     \cos(2\pi f t - \omega) + \sin(2\pi f t - \omega) + b^{(d)}
\end{pmatrix}.
$$

See Figure~\ref{fig:trajectories} b) for an illustration of $\vec{\ell}_{rot}$.

Furthermore, in real-world applications, it is often the case that doctors may either change the settings of the echocardiogram machine during screening or zoom in or out to get better views of specific cardiac structures. 
Additionally, some patients might slightly move while scans are performed, which leads to a heart displacement with respect to the transducer position throughout an echo recording. 
This is particularly prominent in our in-house dataset, which consists of echocardiograms of newborn children. 
Such echocardiograms are not necessarily well represented with a simple periodic trajectory, as over multiple cycles, the spatial structure of a sample shifts and looks different than in the beginning, even though temporal information like the frequency or phase shift is preserved. 
Thus, the current trajectory model fails in such scenarios, which can manifest in two ways: either the model gets stuck in a local optimum with high reconstruction error, or the model tries to reconstruct the video from one long cycle, hence not leveraging the heart cycle periodicity. 
Hence, to account for movements of the recording device, we extend $\vec{\ell}_{rot}$ with a velocity parameter $v\in\mathbb{R}$ that allows the model to learn gradual shifts of the latent trajectory over time, resulting in a trajectory that is no longer circular but a spiral embedded in high dimensional space. 
More formally, we define the spiral trajectory function as
$$
\vec{\ell}_{spiral}(t;f,\omega,v,\vec{b})_i=\vec{\ell}_{rot}(t;f,\omega,\vec{b})_i+tv
$$
See Figure~\ref{fig:trajectories} c) for an illustration of the spiral model.
\subsubsection{Variational Formulation}
Previous work often applied VAEs to anomaly detection, as its generative nature enables more sophisticated variants of reconstruction-based anomaly detection \citep{baur2018deep, chen2020unsupervised,xu2018unsupervised}.
However, the current latent trajectory model is purely deterministic. 
Thus, we introduce the variational latent trajectory model and perform a post-hoc MAP estimation to detect anomalies similar to \cite{chen2020unsupervised}.

We modify the encoder $\phi(X;\theta)$ such that it outputs trajectory parameters $v,f,\omega\in\mathbb{R}$ and $\vec{\mu_b},\vec{\sigma_b}\in\mathbb{R}^d$. 
The model is then extended with a stochastic layer by defining ${\vec{b}\sim q_{\theta}(\vec{b}|X)\coloneqq\mathcal{N}(\vec{\mu_b},diag(\vec{\sigma_b}))}$.
While we aim to learn a distribution over heart shapes, we would also like to accurately identify the frequency $f$, phase shift $\omega$, and spatial shift $v$ given an echo video $X$, instead of sampling them from a latent distribution. 
We thus leave those parameters deterministic.
Next, we define an isotropic Gaussian prior $p(\vec{b}):=\mathcal{N}(0,\mathbb{I})$ on $\vec{b}$ and assume that $x^{(i)}\sim p_{\eta}(X|\vec{b},f,\omega, v)\coloneqq\mathcal{N}(\psi(\vec{\ell}_{spiral}(t^{(i)};f,\omega,v,\vec{b});\eta), \sigma \mathbb{I})$, where $\psi$ is our decoder with weights $\eta$ and $\sigma$ is some fixed constant.
Given these assumptions, we can derive the following \emph{evidence lower bound} (ELBO):
$$
ELBO(X)\coloneqq E_{q_{\theta}(\vec{b}|X)}[\log(p_{\eta}(X|\vec{b},\phi_{f}(X),\phi_{\omega}(X),\phi_{v}(X)))]-KL[q_{\theta}(\vec{b}|X)||p(\vec{b})]
$$ 
Here, $\phi_f(X)$, $\phi_{\omega}(X)$ and $\phi_v(X)$ are the trajectory parameter outputs of the encoder $\phi$ for $f,\omega$ and $v$, respectively.
Note that VAEs on $\vec{\ell}_{circular}$ and $\vec{\ell}_{rot}$ are defined in a similar fashion.
A derivation of this ELBO can be found in Appendix~\ref{appendix:elbo_derivation}.
\subsubsection{Anomaly detection}\label{subsubsec:ad}
The variational formulation of the latent trajectory model allows us to perform anomaly detection by \emph{Maximum a Posteriori} (MAP) inference as proposed in \cite{chen2020unsupervised}.
They suggest to model anomalies as an additive perturbation of a healthy sample.
Following their reasoning we define a healthy sample $X_H\coloneqq (\vec{x}_H^{(j)}, t^{(j)})_{j=1}^T\sim \mathcal{H}$, (anomalous) sample $Y\coloneqq(\vec{y}^{(j)}, t^{(j)})_{j=1}^T\sim\mathcal{D}$, and anomaly perturbation $A\coloneqq (\vec{a}^{(j)}, t^{(j)})_{j=1}^T$, where $\mathcal{H}$ is the healthy data distribution and $\mathcal{D}$ the overall data distribution, and assume that
$$
\vec{y}^{(j)}=\vec{x}^{(j)}_H+\vec{a}^{(j)}
$$
In the case of CHD, $A$ could, e.g., remove walls between heart chambers or produce holes in the myocardium for certain frames.
The anomaly score $\alpha$ can then be defined as $\alpha(Y)\coloneqq \frac{1}{T} \sum_{j=1}^T \|\vec{a}^{(j)}\|^2_2$.
When training a VAE on healthy samples only, i.e. $\vec{a}^{(j)}=0$ for all $j\in \{1,..., T\}$, the variational latent trajectory model learns to approximate $P(X_H)$ by maximizing $ELBO(X_H)$. 
The usual MAP estimation maximizes the posterior distribution of $X_H$ given $Y$. 
By Bayes' theorem
$$
P(X_H|Y)\propto  P(Y|X_H)P(X_H),
$$
the concavity of the logarithm, as well as the fact that $\log(P(X_H))\geq ELBO(X_H)$ it is then possible to estimate $X_H$ by 
$$
\Tilde{X}_H = \argmax_{X_H} (\log(P(Y|X_H))+ELBO(X_H))
$$
To compute the anomaly score we compute $\vec{\Tilde{a}}^{(j)}=\vec{y}^{(j)}-\vec{\Tilde{x}}^{(j)}_H$ and arrive at $\alpha(Y)\coloneqq \frac{1}{T} \sum_{t=1}^T \|\vec{\Tilde{a}}^{(t)}\|^2_2$.
Similar to \cite{chen2020unsupervised}, we choose $\log P(Y|X)=\|(\vec{x}^{(j)}-\vec{y}^{(j)})_{j=1}^T\|_{TV}$, where $\|\cdot\|_{TV}$ denotes the \emph{Total Variation Norm} in $\ell_1$, as this incorporates the assumption that anomalies should consist of contiguous regions rather than single pixel perturbations.
Note that since we have a temporal model, we can incorporate temporal gradients into the TV norm, i.e.
$$
\|X\|_{TV} \coloneqq \sum_{i=1}^{w}\sum_{j=1}^{h}\sum_{k=1}^{T} \|\nabla \vec{x}^{(k)}_{ij}\|_1
$$

In our experiments, we approximate gradients by
$$
\nabla \vec{x}^{(k)}_{ij} \approx \begin{pmatrix}
    \vec{x}^{(k)}_{(i+1)j}-\vec{x}^{(k)}_{(i-1)j} \\[10pt]
    \vec{x}^{(k)}_{i(j+1)}-\vec{x}^{(k)}_{i(j-1)} \\[10pt]
    \vec{x}^{(k+1)}_{ij}-\vec{x}^{(k-1)}_{ij}
\end{pmatrix} 
$$

\section{Cohort}\label{sec:cohort}
\begin{table}[h]
\centering
\caption{Cohort Statistics}
\vspace{4pt}
\begin{tabular}{|l|l|} 
 \hline
 \textbf{Feature}&\textbf{Statistic}\\
 \hline
  No. of Patients &$192$ \\
 \hline
  No. of Patients with no CHD &$69$ \\
 \hline
  No. of Patients with SSD& $5$ \\
 \hline
  No. of Patients with PH & $73$\\
 \hline
  No. of Patients with RVDIL& $73$\\
 \hline
 Age (Days) (Mean$\pm$ SD)& $34\pm 48$\\
 \hline111
 Time until birth (Days) (Mean$\pm$ SD)& $232\pm 46$\\
 \hline
 Weight (Gramms) (Mean$\pm$ SD)& $2774\pm1227$\\
 \hline
 Manufacturer (Ultrasound Machine / Transducer)& GE Logic S8 / S4-10 at 6 MHz\\
 \hline
 Original Video Size (pixel$\times$pixels)&$1440\times866$\\
 \hline
  Video length (frames) (Mean$\pm$ SD)& $122\pm7$\\
 \hline
  Video FPS &$25$ fps\\
 \hline
\end{tabular}
\label{tab:cohort}
\end{table}
The dataset for this study consists of echos of $192$ newborns and infants up to one year of age collected between $2019$ and $2020$ at a single center by a single pediatric cardiologist. 
All examinations were performed with the GE Logic S$8$ ultrasound machine and contain $2$D video sequences of at least $2$ standard echo views, i.e., apical 4-chamber view (4CV) and parasternal long-axis view (PLAX). 
Of the $192$ patients, $123$ suffer from, potentially multiple, CHDs, and $69$ are healthy. See Table~\ref{tab:cohort} for more details.

In order to evaluate anomaly detection performance, a pediatric cardiologist labeled the dataset into three categories. 
These include \emph{Pulmonary Hypertension} (PH), \emph{Right Ventricular Dilation} (RVDil) and \emph{Severe Structural Defects} (SSD).
While PH and RVDil are well-defined pathologies, SSD was defined as a category of multiple rare but severe CHDs, including Ebsteins anomaly, anomalous left coronary artery origin from pulmonary artery (ALCAPA), atrio-ventricular discordance, and ventricular-artery concordance (AVD-VAC), Shone-complex, total anomalous pulmonary venous drainage (TAPVD), tetralogy of Fallot (ToF) and complete atrioventricular septal defect (cAVSD).
We illustrate examples for healthy, SSD, PH, and RVDil echos of both 4CV and PLAX views in Appendix~\ref{appendix:cohort}.

All collected echocardiograms were preprocessed by resizing them to $128\times128$ pixels. 
Additionally, histogram equalization was performed to increase the contrast of the frames, and pixel values were normalized to the range $[0,1]$.
Consequently, models in the experiments in the following section are trained and evaluated on the preprocessed videos.

\section{Experiments} 
In addition to the \emph{variational latent trajectory} (TVAE) model with the \emph{circular} (TVAE-C), \emph{rotated} (TVAE-R) and \emph{spiral} (TVAE-S) trajectories described in Section~\ref{subsec:traj_models}, as a baseline, we train a standard variational autoencoder \citep{kingma2013auto} model on the individual video frames of the dataset.

We run experiments for each of the three CHD categories described in Section~\ref{sec:cohort} by training the models exclusively on samples that do not exhibit these pathologies.
Each experiment is evaluated on $10$ separate data splits, leaving out $30$ healthy patients for evaluation of PH and RVDil and $7$ for SSD, respectively. 
Additionally, every experiment is performed on both the \emph{apical four chamber} (4CV) and \emph{parasternal long axis} (PLAX) views.

\subsection{Implementation Details}
We assume that any anomaly of the heart should always be visible for a certain period over the heart cycle.
It thus suffices to have a model that reconstructs only a fixed number of video frames, as long as at least one heart cycle is present in the video.
The collected videos are recorded with $24$ frames per second (FPS), and we assume that a heart beats at least $30$ times a minute.
Therefore, we decided to subsample the video frequency to $12$ FPS and reconstruct videos with a fixed length of $25$ frames, which is enough to capture at least one cycle in every video.

Reconstructing a fixed number of frames enables us to implement efficient architectures to aggregate echo frames and predict the trajectory parameters.
More specifically, we implement the encoder by concatenating all input frames of the video, hence treating them like different channels of an image, and passing them to a \emph{residual} (\cite{resnets2016}) encoder backbone. 
Each frame $(x^{(i)},t^{(i)})$ is then individually decoded by passing $\vec{\ell}_{circular}(t^{(i)})$, $\vec{\ell}_{rot}(t^{(i)})$ or $\vec{\ell}_{spiral}(t^{(i)})$ to a \emph{deconvolution} (\cite{zeiler2010deconvolutional}) based decoder.
To train the VAE, we used identical encoder and decoder architectures, only changing the first layer to take a single grayscale channel instead of $25$ frames and adapting latent fully connected layers to match dimensions.
For more detailed schematics of the architecture and an overview of the chosen hyperparameters like latent dimension, batch size, or learning rate, we refer to Appendix~\ref{appendix:hparams}.

We pretrained all models on the EchoDynamic dataset to speed up training convergence \citep{ouyang2020video}.
As in most clinical applications, the scarcity of the data makes optimized models prone to overfitting.
To prevent this, we apply data augmentation during training by transforming samples with random affine transformations, brightness adjustments, gamma corrections, blurring, and adding Salt and Pepper noise before performing the forward pass.

\subsection{Reconstruction}\label{sec:reconstruction_exp}
\begin{figure}[h]
\centering
\subfigure[Healthy reconstructions]{
        \includegraphics[width=\textwidth]{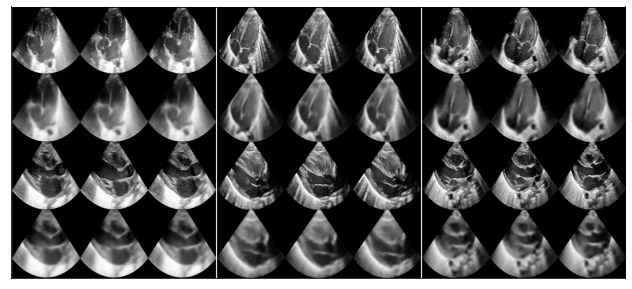}
    }
    \subfigure[SSD reconstructions]{
        \includegraphics[width=\textwidth]{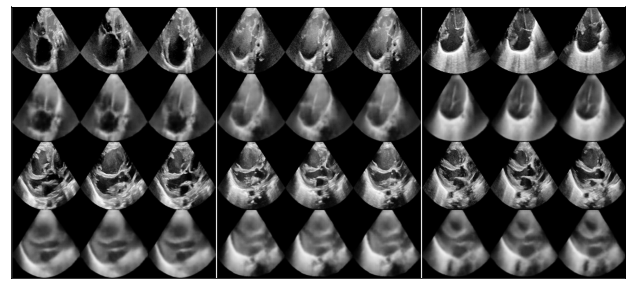}
    }
\caption{Examples of healthy (a) and SSD (b) samples (first and third rows) and their reconstructions (second and fourth rows) using the TVAE-S model.  We sample $3$ frames for each echo from the $25$ frame long sequences. }
\label{fig:ex_recs}
\end{figure}

\begin{table*}[t]
    \centering
    \caption{Apical 4-chamber view reconstruction performance on test data of the proposed approaches (TVAE-C, TVAE-R and TVAE-S) compared with the baseline (VAE). Means and standard deviations are computed across $10$ data splits.}
    \label{tab:rec_scores}
        \centering
    \begin{tabular}{|l|l|c|c|c|c|}
\cline{3-6}
\multicolumn{2}{c|}{}&VAE&TVAE-C&TVAE-R&TVAE-S\\ 
\hline
\multirow{3}{*}{SSD}&MSE&$\boldsymbol{0.013}{\scriptstyle\pm0.0}$&$0.014{\scriptstyle\pm0.0}$&$0.014{\scriptstyle\pm0.0}$&$\boldsymbol{0.013}{\scriptstyle\pm0.0}$\\ 
 \cline{2-6} 
&PSNR&$\boldsymbol{19.008}{\scriptstyle\pm0.11}$&$18.574{\scriptstyle\pm0.15}$&$18.58{\scriptstyle\pm0.13}$&$18.774{\scriptstyle\pm0.09}$\\ 
 \cline{2-6} 
&SSIM&$0.545{\scriptstyle\pm0.01}$&$0.544{\scriptstyle\pm0.01}$&$0.545{\scriptstyle\pm0.01}$&$\boldsymbol{0.552}{\scriptstyle\pm0.01}$\\ 
 \cline{2-6} 
\hline
\multirow{3}{*}{RVDil}&MSE&$\boldsymbol{0.012}{\scriptstyle\pm0.0}$&$0.014{\scriptstyle\pm0.0}$&$0.013{\scriptstyle\pm0.0}$&$0.013{\scriptstyle\pm0.0}$\\ 
 \cline{2-6} 
&PSNR&$\boldsymbol{19.146}{\scriptstyle\pm0.05}$&$18.7{\scriptstyle\pm0.07}$&$18.82{\scriptstyle\pm0.08}$&$18.803{\scriptstyle\pm0.04}$\\ 
 \cline{2-6} 
&SSIM&$\boldsymbol{0.555}{\scriptstyle\pm0.0}$&$0.549{\scriptstyle\pm0.0}$&$0.554{\scriptstyle\pm0.0}$&$\boldsymbol{0.555}{\scriptstyle\pm0.0}$\\ 
 \cline{2-6} 
\hline
\multirow{3}{*}{PH}&MSE&$\boldsymbol{0.012}{\scriptstyle\pm0.0}$&$0.014{\scriptstyle\pm0.0}$&$0.014{\scriptstyle\pm0.0}$&$0.014{\scriptstyle\pm0.0}$\\ 
 \cline{2-6} 
&PSNR&$\boldsymbol{19.084}{\scriptstyle\pm0.07}$&$18.66{\scriptstyle\pm0.07}$&$18.723{\scriptstyle\pm0.08}$&$18.727{\scriptstyle\pm0.07}$\\ 
 \cline{2-6} 
&SSIM&$0.552{\scriptstyle\pm0.0}$&$0.55{\scriptstyle\pm0.0}$&$0.551{\scriptstyle\pm0.0}$&$\boldsymbol{0.553}{\scriptstyle\pm0.0}$\\ 
 \cline{2-6} 
\hline

\end{tabular}
\end{table*}

Reconstruction quality is directly related to reconstruction-based anomaly detection performance, as we rely on the \emph{manifold} and \emph{prototype} assumptions formalized in \cite{ruff2021unifying}.
The manifold assumption is often used in many machine learning-based applications and states that $\mathcal{X}$, the space of healthy echos, can be generated from some latent space $\mathcal{Z}$ by a decoding function $\psi$ and that it is possible to learn a function $\phi$ that encodes $\mathcal{X}$ into $\mathcal{Z}$.
The \emph{better} a learned function $f(x):=\psi(\phi(x))$ reconstructs $x$ on a test set, the better we meet the manifold assumption.
The prototype assumption, on the other hand, assumes that there is some set of prototypes that characterizes the healthy distribution well. 
In our case, the prototypes would be echos corresponding to healthy hearts, i.e., a subset of $\mathcal{X}$.
Under the prototype assumption, our model $f$ must be able to assign a given sample to one of the learned prototypes, i.e., project anomalies to the closest healthy echo.

Table~\ref{tab:rec_scores} contains the scores of the VAE, TVAE-C, TVAE-R, and TVAE-S with respect to the \emph{Mean Squared Error} (MSE), \emph{Peak Signal to Noise Ratio} (PSNR) and \emph{Structural Similarity Index Measure} (SSIM).
We observe how TVAE-C has consistently higher MSE and SSIM errors and lower PSNR than both TVAE-R and TVAE-S. 
Upon inspection of the reconstructed test videos, we notice that, for most seeds, TVAE-C converges to a local optimum where the model learns mean representations of the input videos, thus ignoring the latent dimensions containing temporal information, as described in Section~\ref{sec:methods}.
On the other hand, we did not observe this behavior in TVAE-R and TVAE-S, suggesting that these models indeed capture dynamic properties of echos through the learned latent representations.
Additionally, TVAE-S achieves good echo reconstructions even for samples with transducer position displacement, improving upon TVAE-R and achieving similar performance as VAE despite having a smaller information bottleneck.
The proposed approaches, TVAE-C, TVAE-R, and TVAE-S, encode videos into $d+2$ or $d+3$ trajectory parameters respectively, while the VAE encodes each frame in $\mathbb{R}^{d}$, resulting in a total of $25\times d$ latent parameters.
In conclusion, TVAE-S and the standard VAE fulfill the manifold assumption.
Figure~\ref{fig:ex_recs} presents reconstructed healthy, and SSD samples for the 4CV and PLAX echo views.

\begin{figure}[h]
\centering
\includegraphics[width=\textwidth]{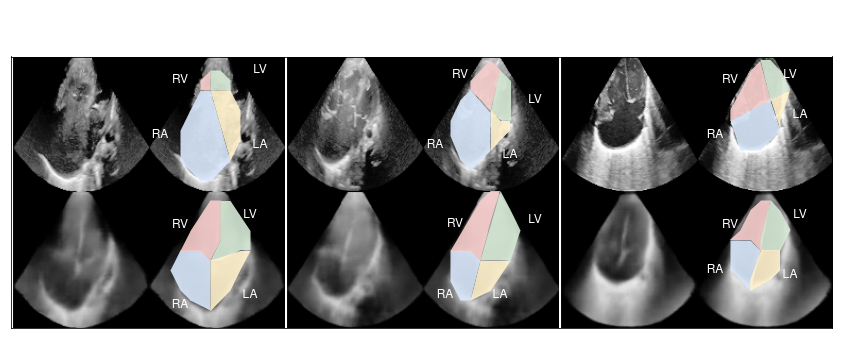}
\caption{Projection of 4CV view anomalous echo (top) to healthy prototype (bottom). Projections of right (R) and left (L) ventricle (V) and atrium (A) are highlighted in color. The reconstruction of SSD samples approximates a healthy version of the input, e.g., by normalizing the scale of the right and left ventricles (left), adding the ventricular septum (middle), or fixing the location of the valves (right).}
\label{fig:ano_proj}
\end{figure}

In Figure~\ref{fig:ano_proj}, we qualitatively demonstrate that TVAE satisfies the prototype assumption.
We observe how the perturbed septum and enlarged/shrunken heart chambers of SSD anomalies are projected to healthy echo reconstructions.

We provide more reconstructions and comprehensive performance comparison of the deterministic and variational models for the 4CV and PLAX echo views in Appendix~\ref{appendix:reconstructions}.

\subsection{Anomaly Detection}
\begin{table*}[t]
\small
    \centering
    \caption{The area under the curve and average precision of the proposed approaches (TVAE-C, TVAE-R, and TVAE-S) compared with the baseline (VAE) on the four-chamber view and long-axis view for the three different CHD labels. Means and standard deviations are computed across $10$ data splits on the test sets. We defined positive labels to correspond to anomalous echos to compute the scores. AP scores of a random classifier are $0.58$ (SSD), $0.72$ (RVDil), and $0.72$ (PH).}
    \label{tab:ad_scores}
        \centering
    \begin{tabular}{|l|l|c|c|c|c|c|c|}
\cline{3-8}
\multicolumn{2}{c|}{}&\multicolumn{2}{c|}{SSD}&\multicolumn{2}{c|}{RVDil}&\multicolumn{2}{c|}{PH} \\ 
\cline{3-8} 
\multicolumn{2}{c|}{}&AUROC&AP&AUROC&AP&AUROC&AP\\ \hline
     \multirow{4}{*}{\smaller{4CV}}&\smaller{VAE}&$0.645{\scriptstyle\pm0.08}$&$0.667{\scriptstyle\pm0.08}$&$0.477{\scriptstyle\pm0.05}$&$0.715{\scriptstyle\pm0.05}$&$0.498{\scriptstyle\pm0.05}$&$0.722{\scriptstyle\pm0.03}$\\ 
 \cline{2-8} 
&\smaller{TVAE-C}&$0.913{\scriptstyle\pm0.09}$&$0.916{\scriptstyle\pm0.11}$&$\boldsymbol{0.6}{\scriptstyle\pm0.05}$&$0.762{\scriptstyle\pm0.04}$&$0.612{\scriptstyle\pm0.05}$&$0.786{\scriptstyle\pm0.04}$\\ 
 \cline{2-8} 
&\smaller{TVAE-R}&$\boldsymbol{0.917}{\scriptstyle\pm0.05}$&$\boldsymbol{0.928}{\scriptstyle\pm0.05}$&$0.594{\scriptstyle\pm0.07}$&$0.771{\scriptstyle\pm0.04}$&$0.629{\scriptstyle\pm0.08}$&$\boldsymbol{0.797}{\scriptstyle\pm0.06}$\\ 
 \cline{2-8} 
&\smaller{TVAE-S}&$0.868{\scriptstyle\pm0.05}$&$0.892{\scriptstyle\pm0.05}$&$0.595{\scriptstyle\pm0.03}$&$\boldsymbol{0.774}{\scriptstyle\pm0.02}$&$\boldsymbol{0.649}{\scriptstyle\pm0.06}$&$0.794{\scriptstyle\pm0.05}$\\ 
 \cline{2-8} 
\hline 
\hline
     \multirow{4}{*}{\smaller{PLAX}}&\smaller{VAE}&$0.628{\scriptstyle\pm0.14}$&$0.457{\scriptstyle\pm0.07}$&$0.455{\scriptstyle\pm0.05}$&$0.702{\scriptstyle\pm0.03}$&$0.432{\scriptstyle\pm0.04}$&$0.695{\scriptstyle\pm0.03}$\\ 
 \cline{2-8} 
&\smaller{TVAE-C}&$0.87{\scriptstyle\pm0.1}$&$0.811{\scriptstyle\pm0.15}$&$0.599{\scriptstyle\pm0.07}$&$\boldsymbol{0.794}{\scriptstyle\pm0.04}$&$0.631{\scriptstyle\pm0.05}$&$0.818{\scriptstyle\pm0.03}$\\ 
 \cline{2-8} 
&\smaller{TVAE-R}&$0.877{\scriptstyle\pm0.08}$&$0.826{\scriptstyle\pm0.1}$&$\boldsymbol{0.61}{\scriptstyle\pm0.04}$&$\boldsymbol{0.794}{\scriptstyle\pm0.02}$&$0.629{\scriptstyle\pm0.06}$&$0.817{\scriptstyle\pm0.03}$\\ 
 \cline{2-8} 
&\smaller{TVAE-S}&$\boldsymbol{0.914}{\scriptstyle\pm0.09}$&$\boldsymbol{0.876}{\scriptstyle\pm0.14}$&$0.592{\scriptstyle\pm0.05}$&$0.791{\scriptstyle\pm0.03}$&$\boldsymbol{0.636}{\scriptstyle\pm0.05}$&$\boldsymbol{0.821}{\scriptstyle\pm0.02}$\\ 
 \cline{2-8} 
\hline

\end{tabular}
\end{table*}

As described in Section~\ref{subsubsec:ad}, we detect anomalies by MAP estimation:
$$
\Tilde{X}_H = \argmax_{X_H} (\log(P(Y|X_H))+ELBO(X_H))))
$$
Due to the reconstruction loss in the ELBO, this optimization problem requires us to backpropagate through the whole model in every step. 
As a result, inference with the standard MAP formulation is inefficient and proved to be infeasible for our experiments.
To circumvent this problem, we assumed the reconstruction part of the ELBO to be constant and solely balanced the posterior with the KL-Divergence of the encoded $\vec{b}$, i.e., how well $X_H$ is mapped to a standard Gaussian, thus computing
$$
\Tilde{X}_H=\argmax_{X_H}(P(Y|X_H)-KL[q(\vec{b}|X_H)||p(\vec{b})])
$$
Solving this optimization procedure results in only backpropagating through the encoder instead of the whole model, which leads to a significant speedup while performance was not affected.

To optimize this objective we initialize $\Tilde{X}_H$ with the reconstructions computed by the respective model, i.e. $\Tilde{X}_H^{(0)}=f(Y)$ for model $f$ and input $Y$.
We then solve the inference problem with the Adam optimizer, incorporating a learning rate of $0.01$ and taking $100$ optimizer steps per sample.
Additionally, we weight the TV norm with a factor of $0.001$.
For each sample $Y$, we define the anomaly score $\alpha(Y)\coloneqq \frac{1}{T} \sum_t^T \|\vec{\Tilde{a}}^{(j)}\|^2_2$ as described in Section~\ref{subsubsec:ad}.
Anomaly detection performance is then evaluated in terms of the \emph{Area Under the Receiver Operator Curve} (AUROC) and \emph{Average Precision} (AP) when considering the anomalies as the positive class.
In Table~\ref{tab:ad_scores}, we provide a complete overview of the results of the anomaly detection experiments over both views.

We observe that the proposed approaches outperform the VAE in all experiments.
This holds especially true when detecting SSD, where our models, TVAE-C, TVAE-R, and TVAE-S, have significantly better performance and can reliably detect such anomalies.
Despite outperforming TVAE-C and TVAE-R in terms of reconstruction quality, we also note that TVAE-S does not always perform better in the anomaly detection task.
We explain the score discrepancies between SSD and RVDil/PH because SSD deviates considerably from the healthy distribution.
RVDil and PH, on the other hand, are more subtle and require expert knowledge and several echocardiogram views to be detected in practice.

Additionally, we argue that we achieve superior performance over VAE since TVAE-R and TVAE-S have considerably smaller latent spaces ($d+2$/$d+3$) than the VAE ($25\cdot d$), but similar performance regarding reconstruction quality as demonstrated in Section~\ref{sec:reconstruction_exp}.
This gives the optimizer more flexibility when solving the MAP problem since the frames of $\Tilde{X}_H$ can be updated independently to encode them on Gaussian parameters close to $\mathcal{N}(0,\mathbb{I})$, which may result in overfitting during MAP estimation.

Another reconstruction-based inference method approach where we simply define $\alpha_f(X)$ over the MSE, i.e. $\alpha_f(X) = \frac{1}{T} \sum_{j=1}^T\|(x^{(j)}-(f(X))^{(j)})\|_2^2$, is presented in Appendix~\ref{appendix:ad_ablation}.

\subsection{Decision Heatmaps}\label{sec:heatmaps}
\begin{figure}[h]
    \centering
    \subfigure[Healthy]{
        \includegraphics[width=0.51\textwidth]{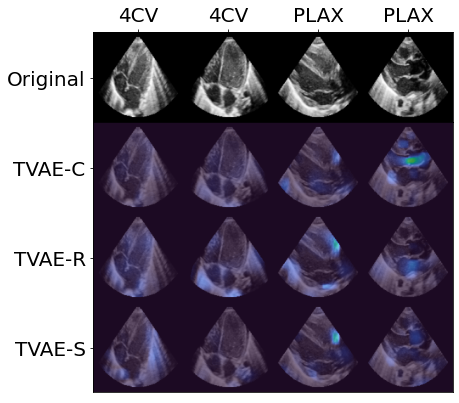}
    }
    \subfigure[Anomalous]{
        \includegraphics[width=0.42\textwidth]{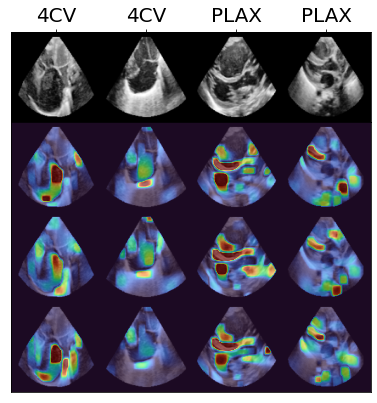}
    }
    \caption{Anomaly response maps of TVAE-R and TVAE-S for healthy samples (a) and echos with CHDs (b). Note how healthy heatmaps are mostly constant, while anomalous maps contain regions with high responses in anomalous regions, corresponding to enlarged ventricles (first/second) or perturbed septums (third/fourth).}
    \label{fig:heatmap}
\end{figure}
This experiment presents how the estimated anomaly perturbation $\Tilde{A}$ can be applied to highlight anomalous regions.
Intuitively, anomalous regions in input echos $Y$ differ more substantially from its healthy projection $X_H$ than healthy regions.
Consequently, this leads to higher magnitude values in the corresponding locations in the frames of $\Tilde{A}$.
In turn, we are able to compute an anomaly heatmap by temporally averaging the estimated anomaly perturbation with $\frac{1}{T}\sum_{j=1}^T\Tilde{\vec{a}}^{(j)}$.
Figure~\ref{fig:heatmap} presents examples of such maps for each TVAE variation.
There we can verify that not only do we have consistently low magnitude responses for healthy echos, but regions corresponding to, e.g., enlarged chambers, are well highlighted for echos with CHDs.
These heatmaps provide TVAE with an additional layer of interpretability and could make our method feasible in a clinical setting, as the reason for decisions made by our method can easily be followed by clinicians. 
This helps practitioners build trust in model decisions and provides a more intuitive explanation of the outputs of our method.
More examples of decision heatmaps are provided in Appendix~\ref{appendix:heatmaps}.

\section{Discussion} 
In this work, we introduce the TVAE; a new generative model designed explicitly for echocardiogram data. 
We propose three variants of the model, TVAE-C, and TVAE-R, which make strong assumptions about the data, and the TVAE-S, which can handle more dynamic inputs.
Throughout this work, we compared the proposed approach to the VAE in terms of its reconstruction performance and anomaly detection capabilities in a new in-house echo dataset consisting of two different echo views of healthy patients and patients suffering from various CHD.
In exhaustive experiments, we demonstrated how TVAE can achieve reconstruction quality comparable to VAE while having a significantly smaller information bottleneck. 
Additionally, we verified that the proposed model can project out-of-distribution samples, i.e., patients suffering from CHD, into the subspace of healthy echos when learning normative priors and concluded that TVAE fulfills crucial assumptions for reconstruction-based anomaly detection.
Consequently, we evaluated the CHD detection performance of our model, where we found that it leads to a considerable improvement over frame-wise VAE with MAP-based anomaly detection.
Furthermore, we demonstrated how TVAE can separate SSD anomalies almost perfectly from healthy echos.
Finally, we present the ability of this model to not only detect but also localize anomalies with heatmaps generated from the MAP output, which could help clinicians with the diagnosis of CHDs.

\paragraph{Limitations and Future Work} Even though we observe convincing results for SSD, performance for the detection of RVDil and PH is still insufficient for clinical application. 
The learned normative prior may not be strong enough for these samples, making it hard to detect them as outliers conclusively.
This is expected given that these defects are rather subtle and our in-house dataset is relatively small.
It would thus be interesting to apply the proposed approach to different and larger cohorts. 
In the future, we plan to collect more samples for our in-house dataset.
With a more extensive dataset, we look forward to exploring methods that would allow combinations of TVAE with one class classification or future frame prediction methods to achieve more robust anomaly detection in echocardiography-based disease detection.

The spiral trajectory of the TVAE-S model assumes continuous movement over the video and might thus still be limiting in situations where sudden movement occurs.
In practice, we did not observe this to be a problem as TVAE-S learned good reconstructions for such samples.
Still, investigating accelerating trajectories could be an exciting direction.
Further, we want to extend the TVAE to multiple modalities such that it is possible to train a model that learns a coherent latent trajectory of multiple echo views of the same heart.
In the future, we are interested in introducing TVAE to modalities in other medical fields by designing trajectory functions that leverage modality-specific characteristics similar to what we did for echos.


\bibliography{refs}

\appendix
\section{Variational Trajectory Model ELBO derivation}\label{appendix:elbo_derivation}
Recall that we define $\vec{b}\sim q_{\theta}(\vec{b}|X)\coloneqq\mathcal{N}(\vec{\mu_b},diag(\vec{\sigma_b}))$ with prior $p(\vec{b})\coloneqq\mathcal{N}(0,\mathbb{I})$, while leaving the other trajectory parameters deterministic.
Note that this effectively means that we define uniform priors $p(f)$, $p(\omega)$ and $p(v)$ over their support, while having posteriors
$$
q_{\theta}(f|X)\coloneqq\delta_{\phi_f(X)}(f),\quad q_{\theta}(\omega|X)\coloneqq\delta_{\phi_{\omega}(X)}(\omega), \quad q_{\theta}(v|X)\coloneqq\delta_{\phi_v(X)}(v)
$$ 
where $\delta_y$ is the Dirac Delta spiking at $y$ and $\phi_f(X)$, $\phi_{\omega}(X)$ and $\phi_v(X)$ are the trajectory parameter outputs of the encoder $\phi$ with weights $\theta$ for $f,\omega$ and $v$ respectively.

Given input sample $x$ and latent $z$, recall that VAEs aim to maximize the \emph{Evidence LOwer Bound} (ELBO):
$$
E_{q_{\theta}(z|x)}[\log(p_{\eta}(x|z))]-KL[q_{\theta}(z|x)||p(z)]
$$
Here, $x$ corresponds to the input echocardiogram $X \coloneqq (\vec{x^{(j)}}, t^{(j)})_{j=1}^T$ whereas $z \coloneqq (\vec{b},f,\omega,v)$.

Note that $\vec{b},f,\omega$ and $v$ are conditionally independent, i.e.
$$
q_{\theta}(\vec{b},f,\omega,v|X)=q_{\theta}(\vec{b}|X)q_{\theta}(f|x)q_{\theta}(\omega|X)q_{\theta}(v|X)
$$
The KL divergence is additive for joint distributions of independent random variables, i.e. for $P=(P_1,P_2)$ and $Q=(Q_1,Q_2)$, where $P_1,P_2,Q_1$ and $Q_2$ are independent, it holds that
$$
KL(P||Q)=KL(P_1||Q_1)+KL(P_2||Q_2)
$$
We can thus rewrite the ELBO as
\begin{align*}
&E_{q_{\theta}(\vec{b},f,\omega,v|X)}[\log(p_{\eta}(X|\vec{b},f,\omega,v))]\\
-&KL[q_{\theta}(\vec{b}|X)||p(\vec{b})]-KL[q_{\theta}(f|X)||p(f)]\\
-&KL[q_{\theta}(\omega|X)||p(\omega)]-KL[q_{\theta}(v|X)||p(v)]
\end{align*}
Since we assumed a uniform prior for $f, \omega$ and $v$, their KL-Divergence terms become constant under the Dirac Delta distribution.
We can thus ignore the respective terms in the ELBO during optimization as they do not change the result of the $argmax$.

Additionally, since
$$
\int\delta_y(x)f(x)dx=f(y)
$$
we can rewrite the ELBOs reconstruction term as
\begin{align*}
    E_{q_{\theta}(\vec{b},f,\omega,v|X)}&[\log(p_{\eta}(X|\vec{b},f,\omega,v))] \\
    &=\int \delta_{\phi_f(X)}(f) \delta_{\phi_{\omega}(X)}(\omega)\delta_{\phi_v(X)}(v)q_{\theta}(\vec{b}|X)\log(p_{\eta}(X|\vec{b},f,\omega,v))d\vec{b}dfd\omega dv\\
    &=\int q_{\theta}(\vec{b}|X)\log(p_{\eta}(X|\vec{b},\phi_{f}(X),\phi_{\omega}(X),\phi_{v}(X)))d\vec{b}\\
    &=E_{q_{\theta}(\vec{b}|X)}[\log(p_{\eta}(X|\vec{b},\phi_{f}(X),\phi_{\omega}(X),\phi_{v}(X)))]
\end{align*}
Finally, this leads to the following reformulation of the ELBO objective:
$$
E_{q_{\theta}(\vec{b}|X)}[\log(p_{\eta}(X|\vec{b},\phi_{f}(X),\phi_{\omega}(X),\phi_{v}(X)))]-KL[q_{\theta}(\vec{b}|X)||p(\vec{b})]
$$
\section{Cohort Examples}\label{appendix:cohort}
\begin{figure}[h]
    \centering
    \includegraphics[width=0.8\textwidth]{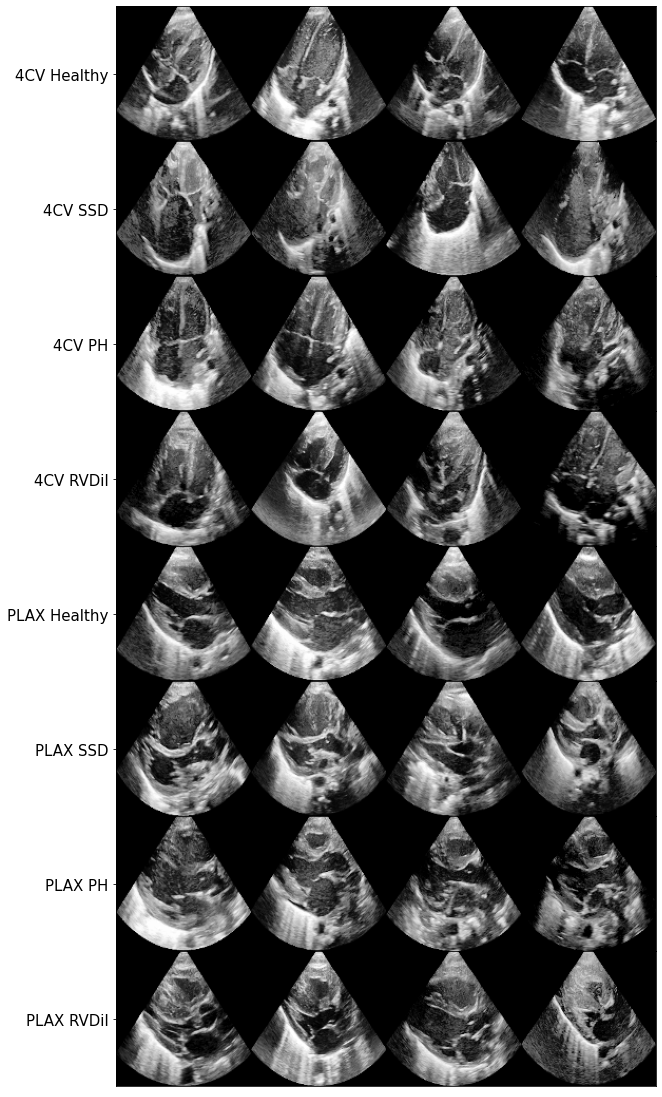}
    \caption{Examples of each label of the cohort in 4CV and PLAX views.}
    \label{fig:cohort_ex}
\end{figure}

To give some intuition on what CHDs look like in different views, we provide examples in Figure~\ref{fig:cohort_ex} and compare them to healthy samples.

\section{Architecture}\label{appendix:hparams}
\begin{table}[h]
\centering
\caption{Hyperparameters chosen across our experiments.}
\label{tab:hparams}
\begin{tabular}{|l || c| c|} 
 \hline
 Hyperparameter&AE/VAE & TAE/TVAE\\
 \hline
 \hline
 Latent Dimension & 64&66/67\\&&(b:64; f:1; $\omega$:1; v:1)\\
 \hline
  Batch Size & 128&64 \\
 \hline
 Steps &5000&106500 \\
 \hline
 Number of Frames &1&25\\
 \hline 
 Optimizer &Adam&Adam\\
 \hline
 Learning Rate&$10^{-4}$&$10^{-4}$\\
 \hline
 Reconstruction Loss & MSE &MSE\\
 \hline
 VAE $\beta$ & 1 & 1\\
 \hline
\end{tabular}
\end{table}
\begin{figure}
    \centering
    \subfigure[Convolution block]{
        \includegraphics[width=0.45\textwidth]{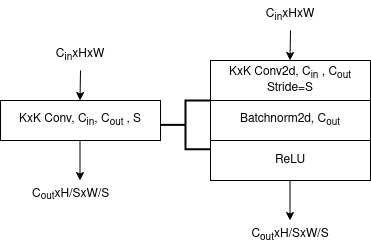}
    }
    \subfigure[Deconvolution block]{
        \includegraphics[width=0.45\textwidth]{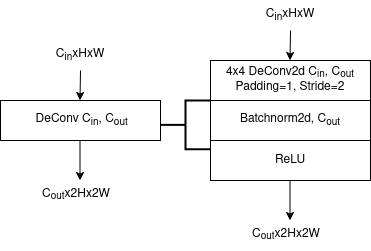}
    }
    \subfigure[Linear block]{
        \includegraphics[width=0.36\textwidth]{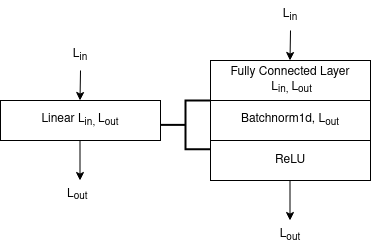}
    }
    \subfigure[Residual block]{
        \includegraphics[width=0.58\textwidth]{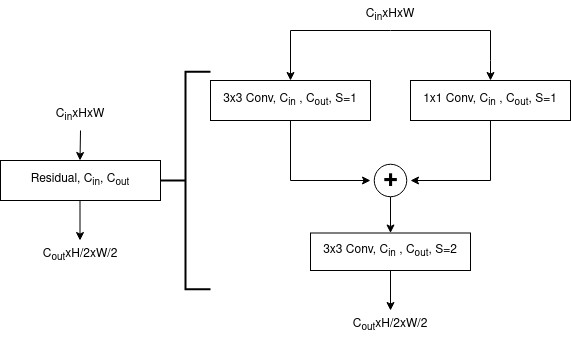}
    }
    \caption{Definitions of the encoder/decoder building blocks.}
    \label{fig:arch_building_blocks}
\end{figure}
\begin{figure}[h]
    \centering
    \subfigure[Encoder]{
        \includegraphics[width=0.2\textwidth]{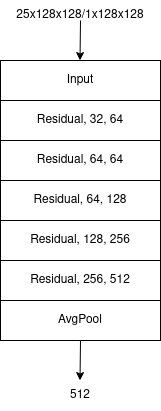}
    }
    \hspace{100pt}
    \subfigure[Decoder]{
        \includegraphics[width=0.2\textwidth]{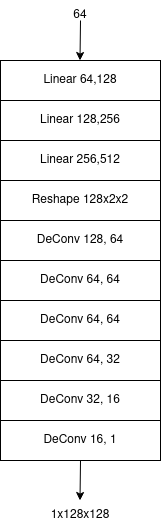}
    }
    
    \subfigure[Latent space components]{
        \includegraphics[width=0.45\textwidth]{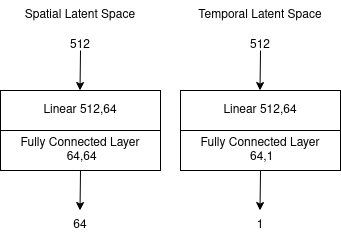}
    }
    \caption{Architectures of encoder (a), decoder (b) and latent space components (c). Spatial latent space components are used to learn $z$, $\mu$ and $\sigma$ for AE/VAE or $b$, $\mu_b$ and $\sigma_b$ for TAE/TVAE. Temporal latent space components learn $f, \omega$ or $v$ for TAE/TVAE.}
    \label{fig:arch}
\end{figure}

We provide schematics for the building blocks of our architectures in Figure~\ref{fig:arch_building_blocks} and describe our experiments' encoder/decoder architecture in Figure~\ref{fig:arch}.

Table~\ref{tab:hparams} contains the hyperparameters used in our experiments.
Except for the number of steps, we kept hyperparameters mostly the same for all models.
This is because, in contrast to the frame-wise models, TAE and TVAE models required many more steps to converge.
We suspect this because the input's dimensionality is $25$ times larger, and the model thus requires more parameter updates to converge to a suitable optimum that results in good reconstructions.
The batch size was chosen according to GPU memory capacity.

\section{Further Reconstruction Experiments}\label{appendix:reconstructions}
In addition to the reconstruction quality experiments provided in Section~\ref{sec:reconstruction_exp}, we compared the performance of the variational models to deterministic ones (i.e., standard autoencoder and non-variational trajectory models). 
As seen in Table~\ref{tab:ablation_rec}, the deterministic trajectory models result in similar performance to the variational models and are even slightly better with respect to the structural similarity score. 
Even though trained on the same architecture and for the same number of steps as the VAE, the autoencoder did not seem to produce very good reconstruction scores in this experiment.
We suspect this may be an artifact of overfitting due to the small training set.

\begin{figure}[h]
    \centering
    \includegraphics[width=\textwidth]{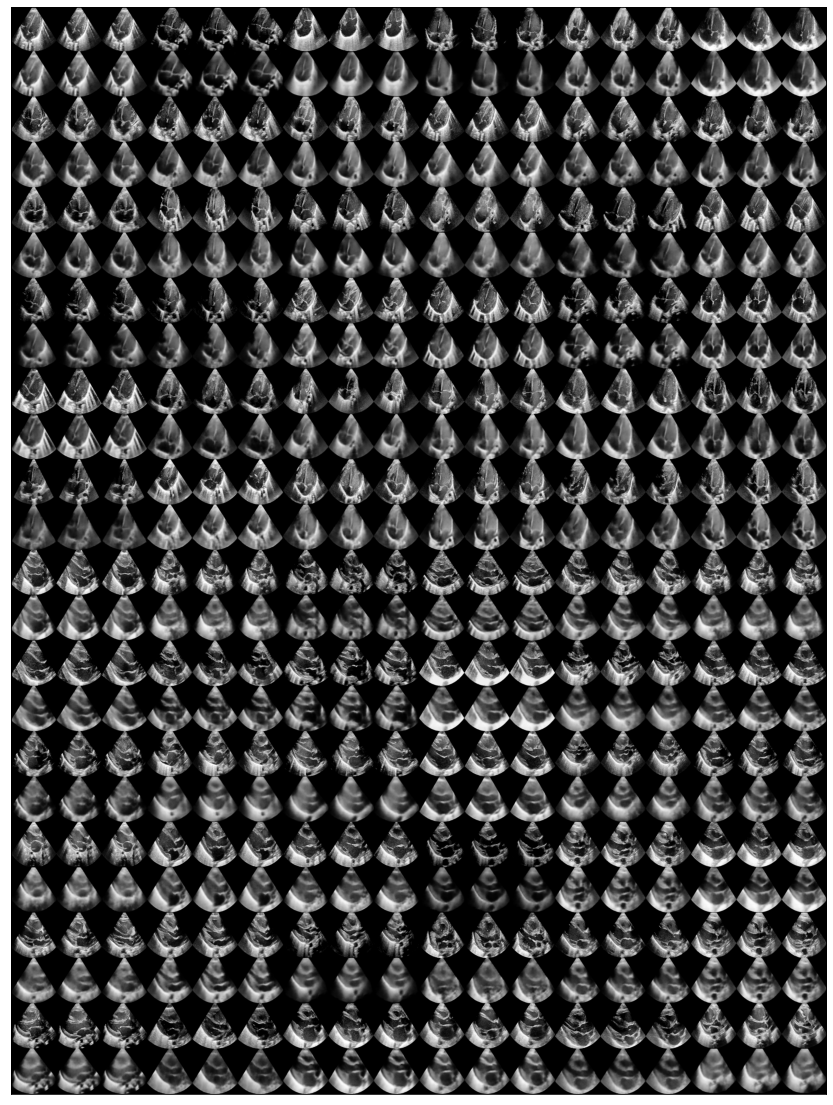}
    \caption{More TVAE-S reconstructions.}
    \label{fig:tvae_s_ablation_recs}
\end{figure}

We provide more reconstructions of TVAE-S in Figure~\ref{fig:tvae_s_ablation_recs}.

\begin{sidewaystable*}[t]
    \centering
    \caption{Reconstruction scores across all introduced models and compared to AE/VAE.}
    \label{tab:ablation_rec}
        \centering
    \begin{tabular}{|l|l|c|c|c|c|c|c|c|c|c|}
\cline{3-11}
\multicolumn{2}{c|}{}&\multicolumn{3}{c|}{SSD}&\multicolumn{3}{c|}{RVDil}&\multicolumn{3}{c|}{PH} \\ 
\cline{3-11} 
\multicolumn{2}{c|}{}&MSE&PSNR&SSIM&MSE&PSNR&SSIM&MSE&PSNR&SSIM\\ 
\hline
     \multirow{8}{*}{\smaller{4CV}}&\smaller{AE}&$0.427{\scriptstyle\pm0.07}$&$4.692{\scriptstyle\pm0.65}$&$0.246{\scriptstyle\pm0.01}$&$0.526{\scriptstyle\pm0.1}$&$3.722{\scriptstyle\pm0.78}$&$0.246{\scriptstyle\pm0.0}$&$0.57{\scriptstyle\pm0.08}$&$3.248{\scriptstyle\pm0.54}$&$0.244{\scriptstyle\pm0.0}$\\ 
 \cline{2-11} 
&\smaller{VAE}&$\boldsymbol{0.013}{\scriptstyle\pm0.0}$&$\boldsymbol{19.02}{\scriptstyle\pm0.11}$&$0.546{\scriptstyle\pm0.01}$&$\boldsymbol{0.012}{\scriptstyle\pm0.0}$&$\boldsymbol{19.146}{\scriptstyle\pm0.05}$&$0.555{\scriptstyle\pm0.0}$&$\boldsymbol{0.012}{\scriptstyle\pm0.0}$&$\boldsymbol{19.093}{\scriptstyle\pm0.07}$&$0.553{\scriptstyle\pm0.0}$\\ 
 \cline{2-11} 

 \cline{2-11} 
&\smaller{TAE-R}&$\boldsymbol{0.013}{\scriptstyle\pm0.0}$&$18.964{\scriptstyle\pm0.15}$&$0.555{\scriptstyle\pm0.01}$&$0.013{\scriptstyle\pm0.0}$&$19.065{\scriptstyle\pm0.05}$&$\boldsymbol{0.563}{\scriptstyle\pm0.0}$&$0.013{\scriptstyle\pm0.0}$&$19.017{\scriptstyle\pm0.07}$&$\boldsymbol{0.562}{\scriptstyle\pm0.0}$\\ 
 \cline{2-11} 
&\smaller{TAE-S}&$\boldsymbol{0.013}{\scriptstyle\pm0.0}$&$18.949{\scriptstyle\pm0.1}$&$\boldsymbol{0.557}{\scriptstyle\pm0.01}$&$0.013{\scriptstyle\pm0.0}$&$18.926{\scriptstyle\pm0.06}$&$0.559{\scriptstyle\pm0.0}$&$0.013{\scriptstyle\pm0.0}$&$18.897{\scriptstyle\pm0.07}$&$0.558{\scriptstyle\pm0.0}$\\ 
 \cline{2-11} 
&\smaller{TVAE-C}&$0.014{\scriptstyle\pm0.0}$&$18.562{\scriptstyle\pm0.15}$&$0.544{\scriptstyle\pm0.01}$&$0.014{\scriptstyle\pm0.0}$&$18.707{\scriptstyle\pm0.07}$&$0.549{\scriptstyle\pm0.0}$&$0.014{\scriptstyle\pm0.0}$&$18.661{\scriptstyle\pm0.07}$&$0.55{\scriptstyle\pm0.0}$\\ 
 \cline{2-11} 
&\smaller{TVAE-R}&$0.014{\scriptstyle\pm0.0}$&$18.568{\scriptstyle\pm0.15}$&$0.545{\scriptstyle\pm0.01}$&$0.013{\scriptstyle\pm0.0}$&$18.813{\scriptstyle\pm0.09}$&$0.554{\scriptstyle\pm0.0}$&$0.014{\scriptstyle\pm0.0}$&$18.73{\scriptstyle\pm0.08}$&$0.552{\scriptstyle\pm0.0}$\\ 
 \cline{2-11} 
&\smaller{TVAE-S}&$\boldsymbol{0.013}{\scriptstyle\pm0.0}$&$18.784{\scriptstyle\pm0.09}$&$0.551{\scriptstyle\pm0.01}$&$0.013{\scriptstyle\pm0.0}$&$18.797{\scriptstyle\pm0.04}$&$0.554{\scriptstyle\pm0.0}$&$0.013{\scriptstyle\pm0.0}$&$18.747{\scriptstyle\pm0.08}$&$0.554{\scriptstyle\pm0.0}$\\ 
 \cline{2-11} 
\hline 
\hline
     \multirow{8}{*}{\smaller{PLAX}}&\smaller{AE}&$0.518{\scriptstyle\pm0.14}$&$3.724{\scriptstyle\pm1.01}$&$0.226{\scriptstyle\pm0.0}$&$0.649{\scriptstyle\pm0.07}$&$2.545{\scriptstyle\pm0.5}$&$0.23{\scriptstyle\pm0.0}$&$0.748{\scriptstyle\pm0.11}$&$1.935{\scriptstyle\pm0.59}$&$0.228{\scriptstyle\pm0.0}$\\ 
 \cline{2-11} 
&\smaller{VAE}&$\boldsymbol{0.017}{\scriptstyle\pm0.0}$&$\boldsymbol{17.874}{\scriptstyle\pm0.15}$&$0.524{\scriptstyle\pm0.0}$&$\boldsymbol{0.015}{\scriptstyle\pm0.0}$&$\boldsymbol{18.394}{\scriptstyle\pm0.05}$&$\boldsymbol{0.541}{\scriptstyle\pm0.0}$&$\boldsymbol{0.015}{\scriptstyle\pm0.0}$&$\boldsymbol{18.442}{\scriptstyle\pm0.03}$&$\boldsymbol{0.542}{\scriptstyle\pm0.0}$\\ 
 \cline{2-11} 
&\smaller{TAE-C}&$0.018{\scriptstyle\pm0.0}$&$17.576{\scriptstyle\pm0.12}$&$0.518{\scriptstyle\pm0.0}$&$0.016{\scriptstyle\pm0.0}$&$17.986{\scriptstyle\pm0.1}$&$0.532{\scriptstyle\pm0.0}$&$0.016{\scriptstyle\pm0.0}$&$18.117{\scriptstyle\pm0.07}$&$0.536{\scriptstyle\pm0.0}$\\ 
 \cline{2-11} 
&\smaller{TAE-R}&$\boldsymbol{0.017}{\scriptstyle\pm0.0}$&$17.836{\scriptstyle\pm0.15}$&$\boldsymbol{0.528}{\scriptstyle\pm0.01}$&$\boldsymbol{0.015}{\scriptstyle\pm0.0}$&$18.196{\scriptstyle\pm0.13}$&$0.54{\scriptstyle\pm0.0}$&$\boldsymbol{0.015}{\scriptstyle\pm0.0}$&$18.192{\scriptstyle\pm0.12}$&$0.54{\scriptstyle\pm0.0}$\\ 
 \cline{2-11} 
&\smaller{TAE-S}&$\boldsymbol{0.017}{\scriptstyle\pm0.0}$&$17.785{\scriptstyle\pm0.12}$&$0.525{\scriptstyle\pm0.0}$&$\boldsymbol{0.015}{\scriptstyle\pm0.0}$&$18.184{\scriptstyle\pm0.04}$&$0.539{\scriptstyle\pm0.0}$&$\boldsymbol{0.015}{\scriptstyle\pm0.0}$&$18.26{\scriptstyle\pm0.04}$&$\boldsymbol{0.542}{\scriptstyle\pm0.0}$\\ 
 \cline{2-11} 
&\smaller{TVAE-C}&$0.019{\scriptstyle\pm0.0}$&$17.264{\scriptstyle\pm0.15}$&$0.509{\scriptstyle\pm0.01}$&$0.017{\scriptstyle\pm0.0}$&$17.795{\scriptstyle\pm0.05}$&$0.526{\scriptstyle\pm0.0}$&$0.017{\scriptstyle\pm0.0}$&$17.778{\scriptstyle\pm0.07}$&$0.526{\scriptstyle\pm0.0}$\\ 
 \cline{2-11} 
&\smaller{TVAE-R}&$0.018{\scriptstyle\pm0.0}$&$17.555{\scriptstyle\pm0.18}$&$0.519{\scriptstyle\pm0.01}$&$0.016{\scriptstyle\pm0.0}$&$17.958{\scriptstyle\pm0.08}$&$0.533{\scriptstyle\pm0.0}$&$0.016{\scriptstyle\pm0.0}$&$17.971{\scriptstyle\pm0.12}$&$0.533{\scriptstyle\pm0.0}$\\ 
 \cline{2-11} 
&\smaller{TVAE-S}&$0.018{\scriptstyle\pm0.0}$&$17.633{\scriptstyle\pm0.09}$&$0.517{\scriptstyle\pm0.0}$&$0.016{\scriptstyle\pm0.0}$&$18.115{\scriptstyle\pm0.1}$&$0.536{\scriptstyle\pm0.0}$&$\boldsymbol{0.015}{\scriptstyle\pm0.0}$&$18.152{\scriptstyle\pm0.08}$&$0.537{\scriptstyle\pm0.0}$\\ 
 \cline{2-11} 
\hline 
\end{tabular}
\end{sidewaystable*}

\section{Reconstruction error-based anomaly detection and one class classification.}\label{appendix:ad_ablation}
A common alternative to MAP-based anomaly detection is the detection of anomalies based on the model's reconstruction error.
This means, for model $f$, sample $x\in\mathcal{X}$ and data space $\mathcal{X}$, we would simply define $\alpha_f(x)=\|x-f(x)\|_2^2$.
In order to quantify the performance of the non-variational dynamic trajectory model (TAE) and to have a comparison to a standard autoencoder trained on single frame reconstruction, we performed another ablation on AE, VAE, and the variants of TAE and TVAE.
As an additional baseline, we also implemented the deep one class classification method introduced in \cite{ruff2018deep}.
We present the results of this ablation in Table~\ref{tab:ad_ablation}.

\begin{table*}[h]
    \centering
    \caption{Area under the curve and average precision for experiments performed with one class classification (OCC) or reconstruction-based with anomaly score $\alpha_f(x)=\frac{1}{T}\sum_{t=1}^T\|x^{(t)}-f^{(t)}(x)\|_2^2$.}
    \label{tab:ad_ablation}
    \begin{tabular}{|l|l|c|c|c|c|c|c|}
\cline{3-8}
\multicolumn{2}{c|}{}&\multicolumn{2}{c|}{SSD}&\multicolumn{2}{c|}{RVDil}&\multicolumn{2}{c|}{PH} \\ 
\cline{3-8} 
\multicolumn{2}{c|}{}&AUROC&AP&AUROC&AP&AUROC&AP\\ \hline
     \multirow{8}{*}{\smaller{4CV}}&\smaller{OCC}&$0.51{\scriptstyle\pm0.05}$&$0.509{\scriptstyle\pm0.04}$&$0.498{\scriptstyle\pm0.01}$&$0.719{\scriptstyle\pm0.01}$&$0.505{\scriptstyle\pm0.01}$&$0.72{\scriptstyle\pm0.01}$\\ 
    \cline{2-8}
&\smaller{AE}&$0.566{\scriptstyle\pm0.1}$&$0.602{\scriptstyle\pm0.09}$&$\boldsymbol{0.634}{\scriptstyle\pm0.05}$&$\boldsymbol{0.816}{\scriptstyle\pm0.03}$&$0.612{\scriptstyle\pm0.03}$&$0.803{\scriptstyle\pm0.02}$\\ 
 \cline{2-8} 
&\smaller{VAE}&$\boldsymbol{0.699}{\scriptstyle\pm0.09}$&$0.732{\scriptstyle\pm0.08}$&$0.619{\scriptstyle\pm0.05}$&$0.803{\scriptstyle\pm0.03}$&$\boldsymbol{0.635}{\scriptstyle\pm0.04}$&$\boldsymbol{0.808}{\scriptstyle\pm0.04}$\\ 
 \cline{2-8} 
&\smaller{TAE-C}&$0.572{\scriptstyle\pm0.09}$&$0.651{\scriptstyle\pm0.05}$&$0.581{\scriptstyle\pm0.04}$&$0.775{\scriptstyle\pm0.03}$&$0.609{\scriptstyle\pm0.02}$&$0.795{\scriptstyle\pm0.01}$\\ 
 \cline{2-8} 
&\smaller{TAE-R}&$0.612{\scriptstyle\pm0.06}$&$0.695{\scriptstyle\pm0.06}$&$0.594{\scriptstyle\pm0.04}$&$0.781{\scriptstyle\pm0.03}$&$0.617{\scriptstyle\pm0.02}$&$0.8{\scriptstyle\pm0.02}$\\ 
 \cline{2-8} 
&\smaller{TAE-S}&$0.558{\scriptstyle\pm0.1}$&$0.646{\scriptstyle\pm0.08}$&$0.612{\scriptstyle\pm0.04}$&$0.794{\scriptstyle\pm0.03}$&$0.614{\scriptstyle\pm0.03}$&$0.802{\scriptstyle\pm0.02}$\\ 
 \cline{2-8} 
&\smaller{TVAE-C}&$0.672{\scriptstyle\pm0.06}$&$0.736{\scriptstyle\pm0.05}$&$0.6{\scriptstyle\pm0.04}$&$0.779{\scriptstyle\pm0.03}$&$0.622{\scriptstyle\pm0.03}$&$0.803{\scriptstyle\pm0.01}$\\ 
 \cline{2-8} 
&\smaller{TVAE-R}&$0.673{\scriptstyle\pm0.07}$&$\boldsymbol{0.745}{\scriptstyle\pm0.07}$&$0.611{\scriptstyle\pm0.03}$&$0.787{\scriptstyle\pm0.02}$&$0.621{\scriptstyle\pm0.04}$&$0.803{\scriptstyle\pm0.03}$\\ 
 \cline{2-8} 
&\smaller{TVAE-S}&$0.616{\scriptstyle\pm0.1}$&$0.679{\scriptstyle\pm0.05}$&$0.61{\scriptstyle\pm0.05}$&$0.786{\scriptstyle\pm0.04}$&$0.631{\scriptstyle\pm0.03}$&$0.8{\scriptstyle\pm0.03}$\\ 
 \cline{2-8} 
\hline 
\hline
     \multirow{8}{*}{\smaller{PLAX}}&\smaller{OCC}&$0.529{\scriptstyle\pm0.05}$&$0.394{\scriptstyle\pm0.07}$&$0.503{\scriptstyle\pm0.01}$&$0.727{\scriptstyle\pm0.01}$&$0.506{\scriptstyle\pm0.01}$&$0.732{\scriptstyle\pm0.01}$\\ 
    \cline{2-8}
     &\smaller{AE}&$0.917{\scriptstyle\pm0.08}$&$0.889{\scriptstyle\pm0.1}$&$0.681{\scriptstyle\pm0.03}$&$0.848{\scriptstyle\pm0.02}$&$0.637{\scriptstyle\pm0.05}$&$\boldsymbol{0.827}{\scriptstyle\pm0.03}$\\ 
 \cline{2-8} 
&\smaller{VAE}&$0.918{\scriptstyle\pm0.07}$&$0.915{\scriptstyle\pm0.05}$&$\boldsymbol{0.683}{\scriptstyle\pm0.03}$&$\boldsymbol{0.851}{\scriptstyle\pm0.02}$&$0.631{\scriptstyle\pm0.05}$&$0.819{\scriptstyle\pm0.03}$\\ 
 \cline{2-8} 
&\smaller{TAE-C}&$0.917{\scriptstyle\pm0.09}$&$0.885{\scriptstyle\pm0.13}$&$0.65{\scriptstyle\pm0.03}$&$0.825{\scriptstyle\pm0.02}$&$\boldsymbol{0.646}{\scriptstyle\pm0.06}$&$\boldsymbol{0.827}{\scriptstyle\pm0.03}$\\ 
 \cline{2-8} 
&\smaller{TAE-R}&$0.913{\scriptstyle\pm0.08}$&$0.878{\scriptstyle\pm0.1}$&$0.68{\scriptstyle\pm0.05}$&$0.84{\scriptstyle\pm0.03}$&$0.639{\scriptstyle\pm0.05}$&$0.823{\scriptstyle\pm0.02}$\\ 
 \cline{2-8} 
&\smaller{TAE-S}&$0.927{\scriptstyle\pm0.07}$&$0.905{\scriptstyle\pm0.09}$&$0.666{\scriptstyle\pm0.03}$&$0.836{\scriptstyle\pm0.01}$&$0.635{\scriptstyle\pm0.04}$&$\boldsymbol{0.827}{\scriptstyle\pm0.02}$\\ 
 \cline{2-8} 
&\smaller{TVAE-C}&$0.927{\scriptstyle\pm0.07}$&$0.907{\scriptstyle\pm0.08}$&$0.65{\scriptstyle\pm0.04}$&$0.828{\scriptstyle\pm0.02}$&$0.617{\scriptstyle\pm0.05}$&$0.814{\scriptstyle\pm0.03}$\\ 
 \cline{2-8} 
&\smaller{TVAE-R}&$0.917{\scriptstyle\pm0.07}$&$0.883{\scriptstyle\pm0.11}$&$0.664{\scriptstyle\pm0.05}$&$0.831{\scriptstyle\pm0.03}$&$0.622{\scriptstyle\pm0.05}$&$0.818{\scriptstyle\pm0.02}$\\ 
 \cline{2-8} 
&\smaller{TVAE-S}&$\boldsymbol{0.935}{\scriptstyle\pm0.08}$&$\boldsymbol{0.916}{\scriptstyle\pm0.11}$&$0.658{\scriptstyle\pm0.05}$&$0.828{\scriptstyle\pm0.02}$&$0.625{\scriptstyle\pm0.04}$&$0.823{\scriptstyle\pm0.02}$\\ 
 \cline{2-8} 
\hline 
\end{tabular}
\end{table*}

\section{More Decision Heatmaps}\label{appendix:heatmaps}
\begin{figure}[h]
    \centering
    \includegraphics[width=\textwidth]{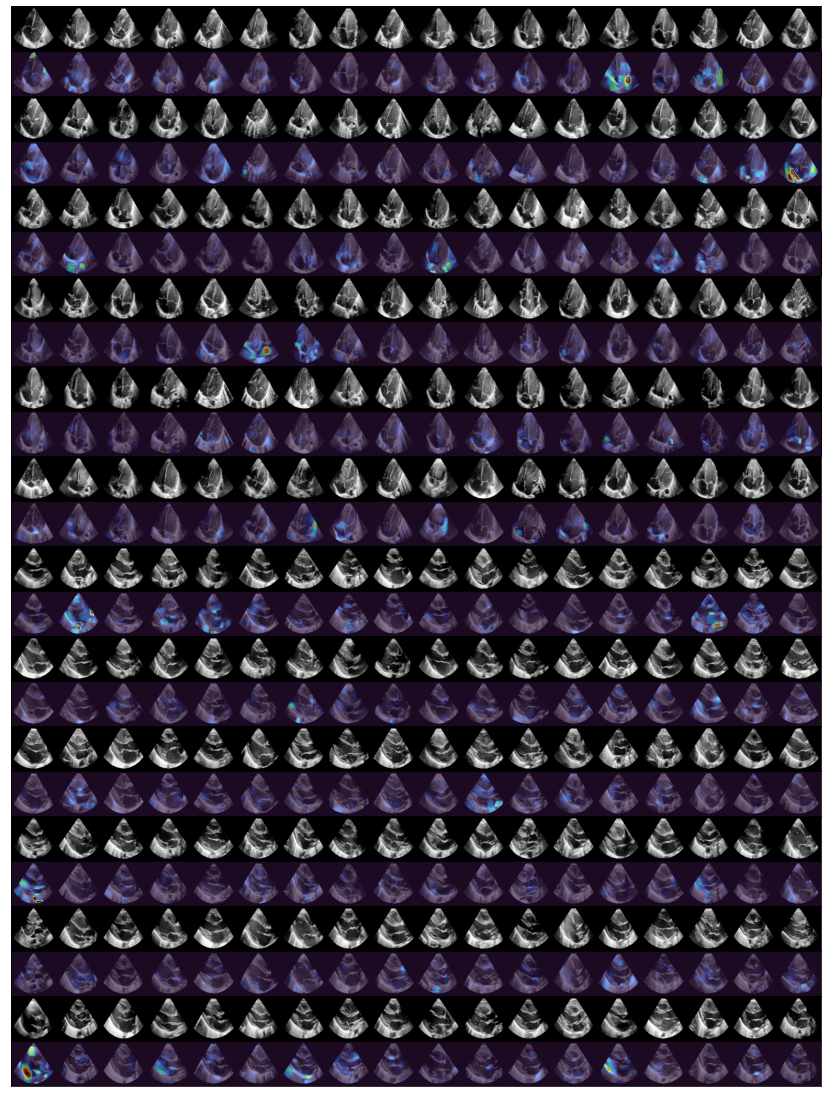}
    \caption{More TVAE-S decision heatmaps for healthy echos.}
    \label{fig:healthy_ablation_heatmaps}
\end{figure}

\begin{figure}[h]
    \centering
    \includegraphics[width=\textwidth]{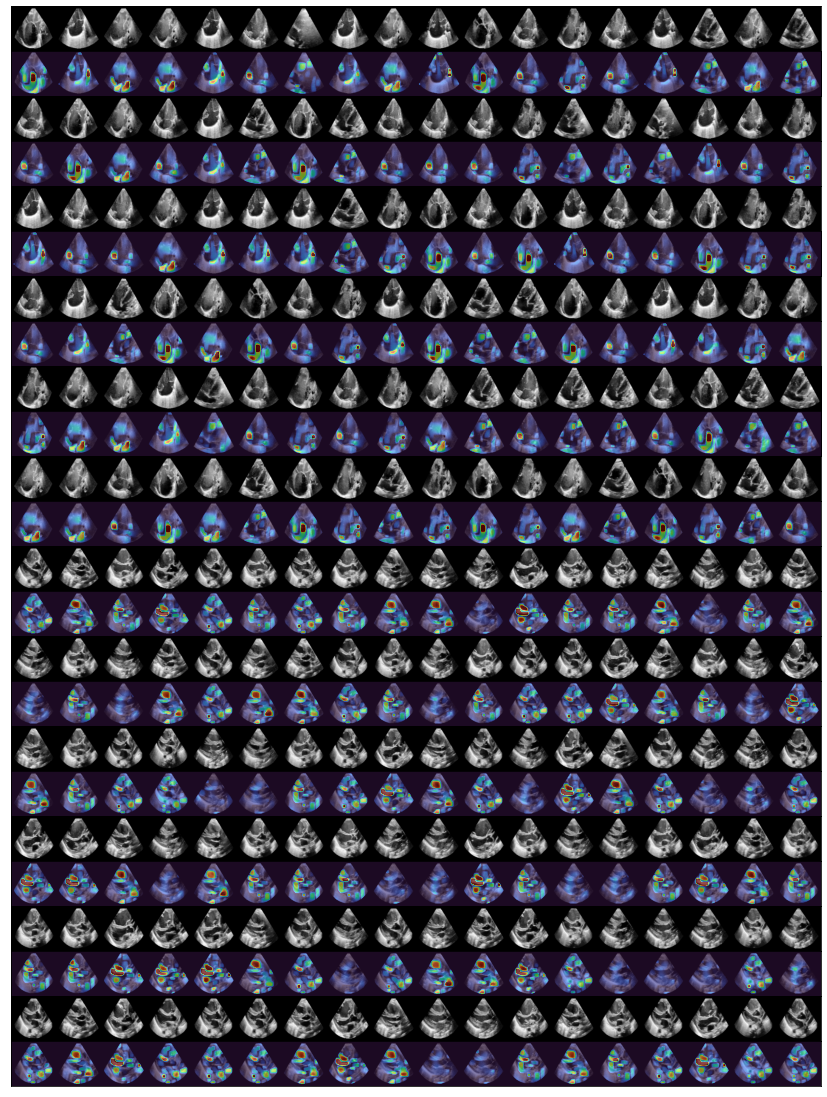}
    \caption{More TVAE-S decision heatmaps for anomalous echos.}
    \label{fig:ano_ablation_heatmaps}
\end{figure}

\begin{figure}[h]
    \centering
    \includegraphics[width=\textwidth]{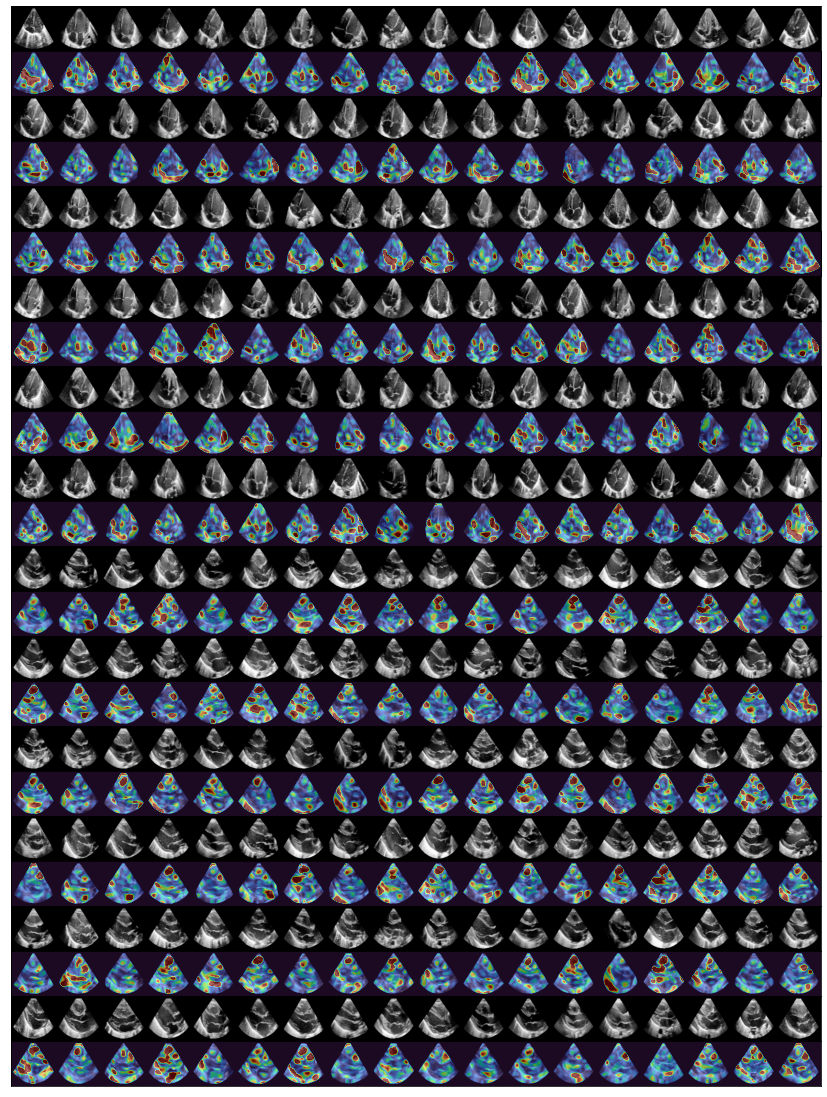}
    \caption{VAE decision heatmaps for healthy echos.}
    \label{fig:healthy_vae_ablation_heatmaps}
\end{figure}

\begin{figure}[h]
    \centering
    \includegraphics[width=\textwidth]{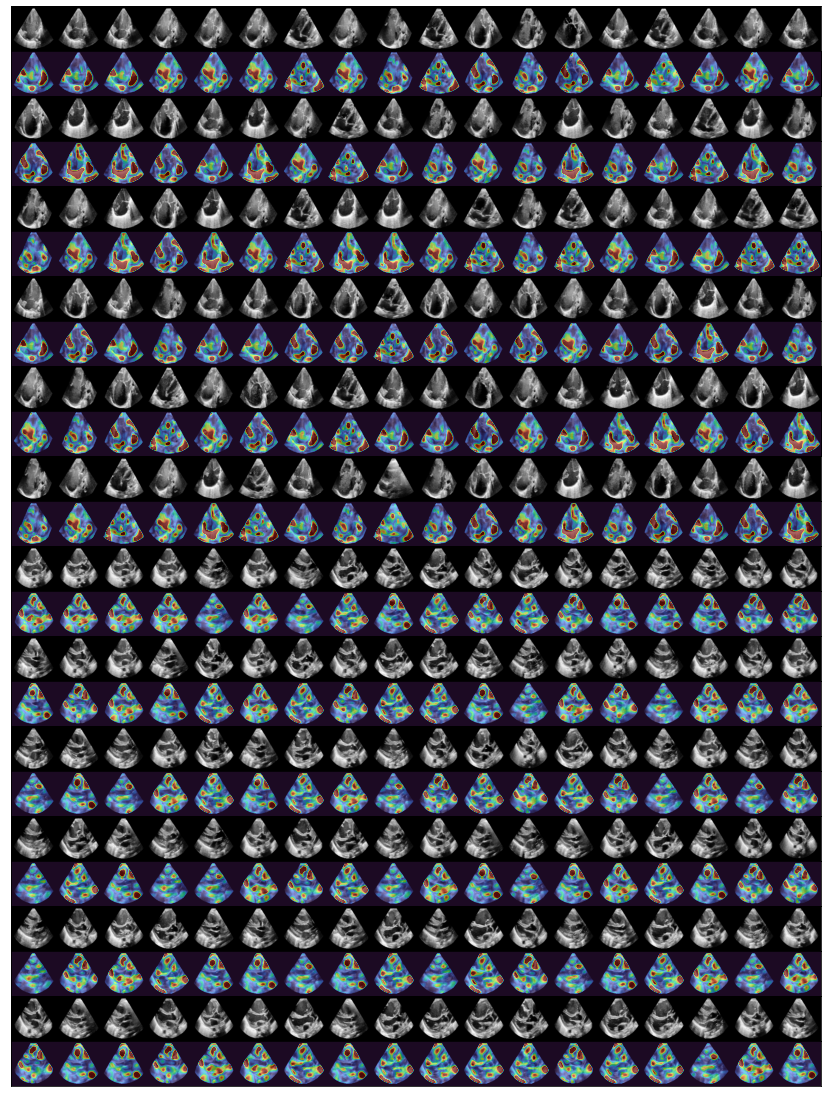}
    \caption{VAE decision heatmaps for anomalous echos.}
    \label{fig:ano_vae_ablation_heatmaps}
\end{figure}

In addition to the heatmaps presented in Section~\ref{sec:heatmaps}, we provide a more extensive collection of TVAE-S decision heatmaps in Figure~\ref{fig:healthy_ablation_heatmaps} and Figure~\ref{fig:ano_ablation_heatmaps} and compare them with heatmaps generated by MAP estimation with a standard VAE \citep{chen2020unsupervised} in Figure~\ref{fig:healthy_vae_ablation_heatmaps} and Figure~\ref{fig:ano_vae_ablation_heatmaps}.
\section{Generated Videos}\label{appendix:generations}
\begin{figure}[h]
    \centering
    \includegraphics[width=\textwidth]{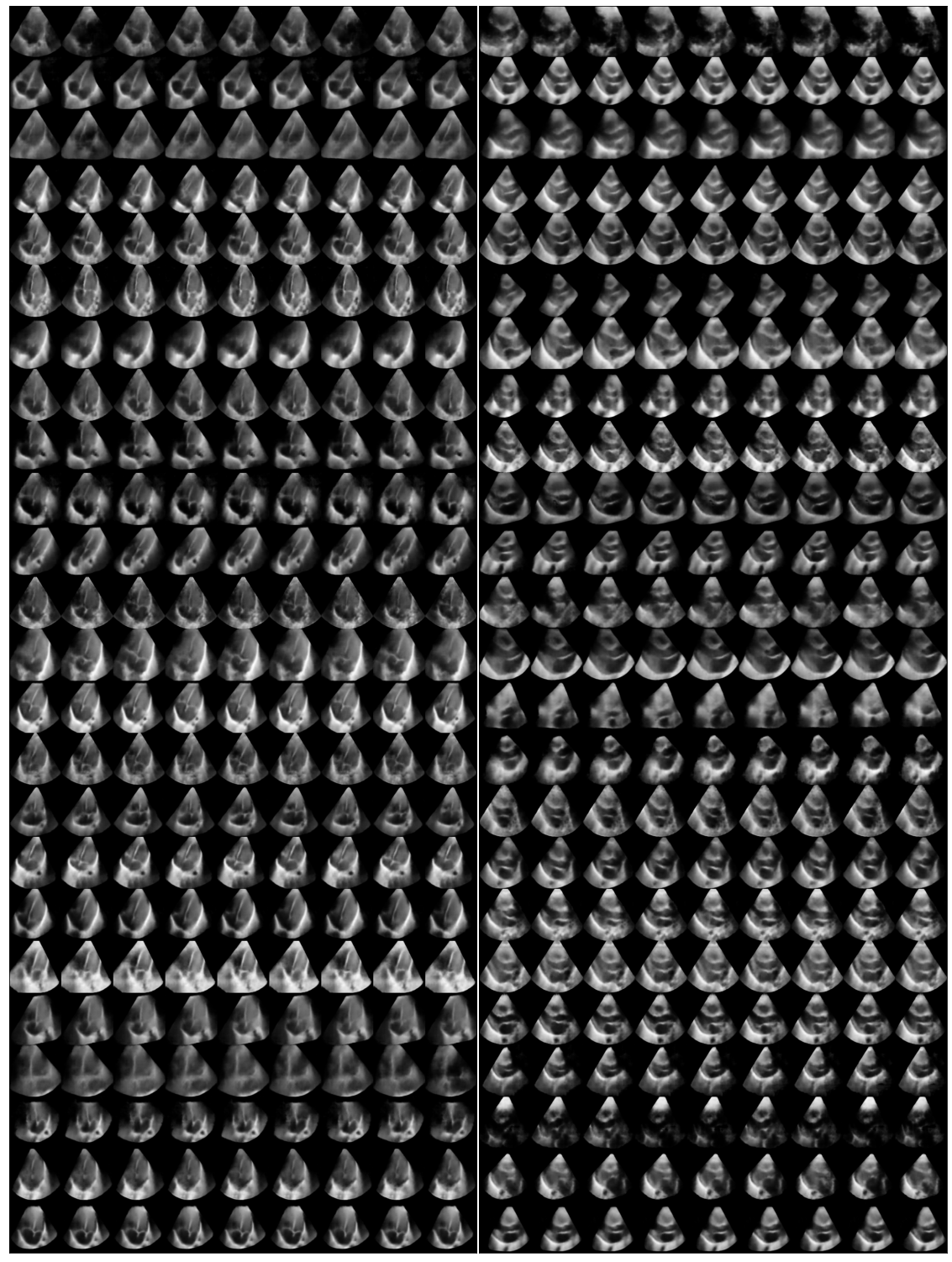}
    \caption{Random TVAE-S generations of samples in 4CV and PLAX views.}
    \label{fig:tvae_s_generations}
\end{figure}

The introduced models TVAE-R and TVAE-S are generative models. 
As such, in addition to producing good reconstructions of existing samples, they allow us to sample from the learned distribution.
To qualitatively validate generative performance, we provide random generations of the TVAE-S model in Figure~\ref{fig:tvae_s_generations} for both 4CV and PLAX views.
\section{Robustness to Domain Shifts}
The proposed approach learns a normative prior on the distribution of the training dataset. 
Further, it assumes that every outlier of the learned distribution corresponds to an anomaly. 
Therefore, a trained model is not robust to domain shifts but needs some finetuning on a new dataset to incorporate the new notion of normality into its normative prior. 
We demonstrate this by combining 100 samples of EchoDynamic, which contains 4CV echocardiograms, with the 4CV views of our in-house dataset for training. 
We assume that EchoDynamic samples correspond to the healthy class. 
EchoDynamic exhibits a considerable distribution shift, as these echocardiograms were recorded by a different machine and collected from an adult population. 
Table~\ref{table:domain_shift} contains the results of this experiment. 
From these results, we can conclude that, for PH and RVDil, learning a new normative prior on this extended dataset does not change the outcome. 
On the other hand, scores for SSD appear less robust, though they still outperform PH and RVDil in AUROC.
\begin{table*}[h]
    \centering
    \caption{Results of the domain shift experiments. We retrained TVAE-S on a modified dataset that combined $100$ EchoDynamic samples with our dataset. We aggregated scores across the test sets of $10$ different data folds. To compute AUROC and AP, we assigned the positive label to the anomalous samples. AP of a random classifier is  $0.085$ (SSD), $0.374$ (RVDil), and $0.374$ (PH) respectively.}\label{table:domain_shift}
    \begin{tabular}{|c|c|c|c|c|c|}
\hline
\multicolumn{2}{|c|}{SSD}&\multicolumn{2}{c|}{RVDil}&\multicolumn{2}{c|}{PH} \\ 
\hline 
AUROC&AP&AUROC&AP&AUROC&AP\\ 
 \hline 
$0.688{\scriptstyle\pm0.09}$&$0.271{\scriptstyle\pm0.06}$&$0.594{\scriptstyle\pm0.04}$&$0.638{\scriptstyle\pm0.03}$&$0.61{\scriptstyle\pm0.02}$&$0.664{\scriptstyle\pm0.02}$\\ 
 \hline
 \end{tabular}
\end{table*}

\section{Anomaly oversensitivity}
To quantify oversensitivity towards anomalies, we can redefine the task of anomaly detection to healthy instance detection, i.e., assigning the positive label to the healthy instances.
Due to its symmetry, the AUROC scores will not change under this new setting; we thus only report AP. 
Table~\ref{table:healthy_instance_detection} contains the results of these experiments.
In the case of SSD, the AP is stable whereas in the case of PH and RVDil, the APs drop. With now $30$ positive versus $73$ negative labels, this is expected to happen. 
Still, results seem to agree with anomaly detection results, as AP scores are still considerably better than random (AP=$0.32$), suggesting that we are not overly sensitive to anomalies.
\begin{table*}[h]
    \centering
    \caption{Average precision of the proposed approaches (TVAE-C, TVAE-R, and TVAE-S) compared to the baseline (VAE) on the four-chamber view and long-axis view for the three different CHD labels. Means and standard deviations are computed across $10$ data splits on the test sets. We defined positive labels to correspond to healthy instances to compute the scores. AP scores of a random classifier are $0.535$ (SSD), $0.317$ (RVDil), and $0.317$ (PH).}\label{table:healthy_instance_detection}
    \begin{tabular}{|l|l|c|c|c|}
\cline{3-5}
\multicolumn{2}{c|}{}&SSD&RVDil&PH \\ 
\cline{3-5} 
\hline
     \multirow{4}{*}{\smaller{4CV}}&\smaller{VAE}&$0.66{\scriptstyle\pm0.1}$&$0.273{\scriptstyle\pm0.04}$&$0.315{\scriptstyle\pm0.06}$\\ 
 \cline{2-5} 
&\smaller{TVAE-C}&$0.91{\scriptstyle\pm0.1}$&$0.419{\scriptstyle\pm0.05}$&$0.392{\scriptstyle\pm0.03}$\\ 
 \cline{2-5} 
&\smaller{TVAE-R}&$\boldsymbol{0.919}{\scriptstyle\pm0.05}$&$\boldsymbol{0.423}{\scriptstyle\pm0.05}$&$0.421{\scriptstyle\pm0.05}$\\ 
 \cline{2-5} 
&\smaller{TVAE-S}&$0.863{\scriptstyle\pm0.07}$&$0.39{\scriptstyle\pm0.05}$&$\boldsymbol{0.458}{\scriptstyle\pm0.07}$\\ 
 \cline{2-5} 
\hline 
\hline
     \multirow{4}{*}{\smaller{PLAX}}&\smaller{VAE}&$0.819{\scriptstyle\pm0.09}$&$0.269{\scriptstyle\pm0.03}$&$0.275{\scriptstyle\pm0.03}$\\ 
 \cline{2-5} 
&\smaller{TVAE-C}&$0.921{\scriptstyle\pm0.07}$&$0.388{\scriptstyle\pm0.07}$&$0.381{\scriptstyle\pm0.05}$\\ 
 \cline{2-5} 
&\smaller{TVAE-R}&$0.927{\scriptstyle\pm0.05}$&$0.393{\scriptstyle\pm0.04}$&$0.387{\scriptstyle\pm0.07}$\\ 
 \cline{2-5} 
&\smaller{TVAE-S}&$\boldsymbol{0.937}{\scriptstyle\pm0.08}$&$\boldsymbol{0.413}{\scriptstyle\pm0.08}$&$\boldsymbol{0.392}{\scriptstyle\pm0.06}$\\ 
 \cline{2-5} 
\hline
\end{tabular}
\end{table*}
\end{document}